\documentclass{article}

\PassOptionsToPackage{numbers, compress}{natbib}
\usepackage[final]{neurips_2024}

\usepackage[utf8]{inputenc} % allow utf-8 input
\usepackage[T1]{fontenc}    % use 8-bit T1 fonts
\usepackage{hyperref}       % hyperlinks
\usepackage{url}            % simple URL typesetting
\usepackage{booktabs}       % professional-quality tables
\usepackage{amsfonts}       % blackboard math symbols
\usepackage{nicefrac}       % compact symbols for 1/2, etc.
\usepackage{microtype}      % microtypography
\usepackage{xcolor}         % colors
\usepackage{appendix}
\usepackage{multirow}
\usepackage{threeparttable}
\usepackage{amsmath}

\usepackage{graphicx}
\usepackage{float}

\newfloat{figtab}{htb}{fgtb}
\makeatletter
  \newcommand\figcaption{\def\@captype{figure}\caption}
  \newcommand\tabcaption{\def\@captype{table}\caption}
\makeatother

\usepackage{amssymb}
\usepackage{bbding}
\usepackage{pifont}
\usepackage{algorithm}
\usepackage{algorithmic}
\usepackage{ragged2e}
\usepackage{booktabs}
\usepackage{color}
\usepackage{multirow}
\usepackage{bm}
\usepackage{bbm}
\usepackage{subfig}
\usepackage{wrapfig}
\usepackage{lipsum}
\def\ie{{\em i.e.}}
\def\eg{{\em e.g.}}
\def\etal{{\em et al.}}

\newcommand{\figref}[1]{Fig.~\ref{#1}}

\usepackage{colortbl}
\definecolor{first}{RGB}{255, 204, 201}
\definecolor{second}{RGB}{255, 228, 207}
\definecolor{third}{RGB}{255, 255, 212}

\definecolor{diffusion_prior}{RGB}{0, 51, 204}
\definecolor{rectified_distribution}{RGB}{0, 128, 43}
\definecolor{render_distribution}{RGB}{204, 0, 0}
\definecolor{inline_prior}{RGB}{198, 95, 16}

\title{How to Use Diffusion Priors under Sparse Views?}

\author{
  Qisen~Wang, \quad Yifan~Zhao\thanks{Correspondence should be addressed to Yifan Zhao and Jia Li. Website: \url{https://cvteam.buaa.edu.cn}}, \quad Jiawei~Ma, \quad Jia~Li$^*$ \\
  State Key Laboratory of Virtual Reality Technology and Systems, SCSE\\
  Beihang University \\
  \texttt{\{wangqisen, zhaoyf, majiawei, jiali\}@buaa.edu.cn} \\
}

\begin{document}

\maketitle

\begin{abstract}
Novel view synthesis under sparse views has been a long-term important challenge in 3D reconstruction. Existing works mainly rely on introducing external semantic or depth priors to supervise the optimization of 3D representations. However, the diffusion model, as an external prior that can directly provide visual supervision, has always underperformed in sparse-view 3D reconstruction using Score Distillation Sampling (SDS) due to the low information entropy of sparse views compared to text, leading to optimization challenges caused by mode deviation. To this end, we present a thorough analysis of SDS from the mode-seeking perspective and propose Inline Prior Guided Score Matching (IPSM), which leverages visual inline priors provided by pose relationships between viewpoints to rectify the rendered image distribution and decomposes the original optimization objective of SDS, thereby offering effective diffusion visual guidance without any fine-tuning or pre-training. Furthermore, we propose the IPSM-Gaussian pipeline, which adopts 3D Gaussian Splatting as the backbone and supplements depth and geometry consistency regularization based on IPSM to further improve inline priors and rectified distribution. Experimental results on different public datasets show that our method achieves state-of-the-art reconstruction quality. The code is released at \url{https://github.com/iCVTEAM/IPSM}.
\end{abstract}

\section{Introduction}
\label{sec:intro}

Novel View Synthesis (NVS) \cite{nerf, 3dgs}, \eg Neural Radiance Fields (NeRF) \cite{nerf, nerf_survey_1, nerf_survey_2} and recently emerged 3D Gaussian Splatting (3DGS) \cite{3dgs, 3dgs_survey_1, 3dgs_survey_2}, requires dense training viewpoints for optimization, as demonstrated in prevailing works \cite{regnerf, freenerf, dngaussian}. Indeed, NVS under sparse views has been an important and challenging task \cite{dietnerf, ds-nerf, regnerf}. Due to the scarcity of viewpoints, most methods of 3D representation reconstruction often fall into over-fitting with sparse views, and cannot synthesize satisfactory novel views \cite{dngaussian, freenerf, sparsenerf}. To address the optimization over-fitting problem under the sparse-view condition, current methods introduce external priors to supervise the optimization of reconstruction like CLIP \cite{clip} semantic information \cite{regnerf}, monocular depth \cite{ds-nerf, dngaussian}, and diffusion visual priors \cite{sparsegs, reconfusion, deceptive-nerf}. 
However, although the diffusion model \cite{diffusion2015, ddpm, sde, analytic_dpm, ddim} as an external prior can provide stronger visual supervision than semantic and depth information, 
it often requires a significant amount of computational resources for \textit{fine-tuning the diffusion prior} \cite{deceptive-nerf} or \textit{pre-training encoders} \cite{reconfusion} with external data. A few works have no fine-tuning and pre-training, but it is difficult to straightly extract diffusion prior knowledge to effectively supplement the missing visual information of sparse views \cite{sparsegs}.

\begin{figure}[t]
    \centering 
    \includegraphics[width=1.0\linewidth]{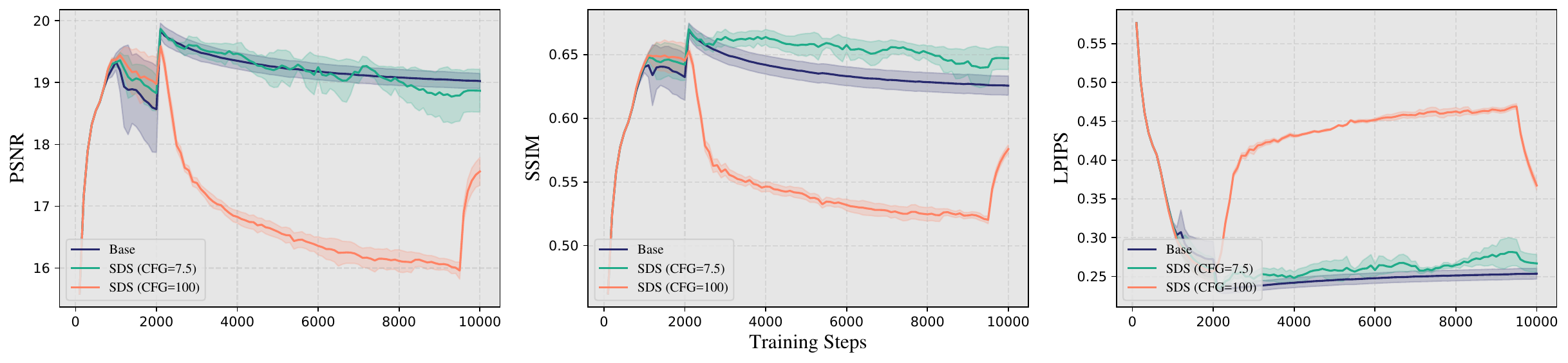} 
    \caption{\textbf{Dilemma of SDS}. Average PSNR$\uparrow$, SSIM$\uparrow$, and LPIPS$\downarrow$ of each iteration on the LLFF test dataset \cite{llff} with Base (without SDS), SDS (CFG=7.5), and SDS (CFG=100). The prior-added period starts from the 2K iteration and ends at the 9.5K iteration. The opacity is also reset at 2K. The details and final training results of SDS are shown in Sec. \ref{sec:comp_sds}.} 
    \label{fig:sds_trend}
\end{figure}

Interestingly, although the diffusion model shows great potential in 3D generation tasks, \eg text-to-3D \cite{sds}, which benefit from the recent rapid development of score distillation techniques \cite{sds, magic123, vsd, luciddreamer}, Score Distillation Sampling (SDS) \cite{sds} shows little visual information guidance ability of the diffusion prior under sparse views and even takes an inhibitory effect on the baseline performance when the input views increase, as shown in \figref{fig:sds_trend}.
\textbf{The SDS dilemma} highlights that score distillation exhibits distinctive optimization characteristics across sparse input views. Consequently, SDS is NOT readily applicable for lifting visual supervision from diffusion priors under sparse views.

With the curiosity of the SDS dilemma in our mind, it can be recognized that the difference between \textit{sparse views} and \textit{text prompts} lies in the inline constraints sparse views bring. For the unsupervised invisible views, unlike text prompts, the ideal rendered image supervision information is not completely absent. Due to the consistency of the 3D geometry and structure, the information exists in the given sparse views, which we refer to the \textbf{inline priors}.
Some researchers \cite{reconfusion} attempt to implicitly encode the given input sparse views to guide the sampling trajectory of the diffusion model, thereby introducing inline priors. Nonetheless, owing to domain shifts between specific scenes and the diffusion prior, a significant amount of external 3D annotated data and computational resources are frequently necessitated for domain rectification \cite{reconfusion}. To this end, a potentially viable approach is exploring the feasibility of adjusting the optimization objective of SDS by incorporating inline priors to facilitate efficient domain rectification without fine-tuning and pre-training.

In this paper, we conduct a comprehensive analysis of SDS from the perspective of mode-seeking. Intuitively, the optimization objective of SDS is to align the rendered image distribution with the target mode in the diffusion prior. However, due to the inherent suboptimality of the rendered image distribution under sparse views, SDS tends to deviate from the target mode, resulting in the SDS dilemma. To tackle this challenge, we present Inline Prior Guided Score Matching (IPSM), a method that rectifies the rendered image distribution by utilizing inline priors.
IPSM leverages the rectified distribution to divide the optimization objective of SDS into two sub-objectives. The rectified distribution, as an intermediate state of the optimization objective, plays a role in controlling the mode-seeking direction, thereby suppressing mode deviation and promoting improvements in reconstruction.
Moreover, we propose the pipeline IPSM-Gaussian, which combines IPSM with the efficient explicit 3D representation 3DGS for sparse-view 3D reconstruction. In addition to IPSM, IPSM-Gaussian integrates depth regularization to support inline priors and geometric consistency regularization to narrow the discrepancy between the rendered image distribution and the rectified distribution at the pixel level. Experimental results demonstrate that IPSM effectively leverages visual knowledge from the diffusion priors to improve sparse-view 3D reconstruction. The presented method achieves superior performance on publicly available datasets.

Overall, our contributions can be summarized as:

\begin{itemize}
    \item \textit{Analysis of SDS from mode-seeking perspective}. We present a comprehensive analysis of SDS optimization characteristics under sparse views, revealing that the mode deviation of SDS results in the optimization dilemma.
    \item \textit{Rectified score distillation method for sparse views}. We propose Inline Prior Guided Score Matching (IPSM), which utilizes inline priors provided by sparse views to rectify rendered image distribution for controlling the direction of seeking the target mode.
    \item \textit{Pipeline using IPSM based on 3DGS}. We present IPSM-Gaussian, a pipeline for sparse-view 3D reconstruction, which adopts IPSM for diffusion guidance, as well as depth and geometry regularization to boost the performance of IPSM. The experiments show that IPSM-Gaussian achieves state-of-the-art reconstruction quality on public datasets.
\end{itemize}

\section{Related Works}
\label{related_works}

\textbf{Novel View Synthesis}.
Novel View Synthesis \cite{nvs1997, nerf, 3dgs, instant-ngp, plenoxels, tensorf, d-nerf} aims to synthesize invisible novel views given a set of images at seen viewpoints while preserving the geometric structure and appearance of the original 3D scene \cite{nerf_in_wild, block-nerf, ref-nerf, mipnerf360, barf, nsvf}. NeRF \cite{nerf, nerf_survey_1, nerf_survey_2}, as an implicit 3D representation, adopts volume rendering to establish an implicit mapping relationship from the positions and ray directions to colors using a Multi-Layer Perception (MLP). Although NeRF can achieve photographic-realistic rendering quality compared to traditional methods, its required training time and rendering speed are not satisfactory \cite{nerf, instant-ngp}. Recently, 3DGS \cite{3dgs, 3dgs_survey_1, 3dgs_survey_2} has garnered attention from researchers by achieving high training speeds and real-time rendering capabilities through explicit modeling of 3D scenes using Gaussian point clouds and rasterization rendering \cite{driving_gs, splatting_avatar, animatable_gs, hifi4g, dreamgaussian}. To this end, we choose 3DGS instead of NeRF as the backbone of 3D representations and adopt it in subsequent experiments.

\textbf{Sparse-view Novel View Synthesis}.
Although current training-based NVS techniques, \ie NeRF \cite{nerf} and 3DGS \cite{3dgs}, can achieve satisfactory rendering quality in scenarios with dense input views, the quality of novel view synthesis significantly decreases under sparse views due to overfitting \cite{flipnerf, sparsenerf, darf, regnerf, ds-nerf, simplenerf}. To tackle this challenge, Yang \etal \cite{freenerf} leverage the optimization properties of MLP and employ annealing strategies for positional encoding \cite{barf} tailored to the characteristics of NeRF, but this cannot be directly applied to 3DGS. More broadly, some works \cite{geconerf, geoaug} leverage the intrinsic relationships between sparse views to augment the data required for model optimization, but this does not address the established condition of information deficiency. More works involve introducing external pre-trained priors as optimization guidance to supervise sparse-view 3D reconstruction. Jain \etal \cite{dietnerf} introduce CLIP \cite{clip} to provide semantic guidance. Li \etal \cite{dngaussian} propose global-local depth regularization with DPT \cite{dpt} for geometric structure guidance. However, the aforementioned prior information cannot directly provide visual supervision for sparse-view NVS like diffusion priors.

\textbf{Sparse-view Novel View Synthesis with Diffusion Priors}.
Although diffusion priors can provide more direct visual guidance, current works are limited by the mode deviation with using diffusion priors directly. Liu \etal \cite{deceptive-nerf} leverage diffusion models to progressively generate pseudo-observations at unseen views. Wu \etal \cite{reconfusion} use PixelNeRF \cite{pixelnerf} to encode sparse inputs for guiding the trajectory of diffusion priors. Unlike score distillation techniques, these works either require fine-tuning the diffusion model for narrowing the mode range \cite{deceptive-nerf}, or pre-training image encoders for guiding the direction of the target mode \cite{reconfusion}, both of which consume many resources \cite{deceptive-nerf, reconfusion}. Xiong \etal \cite{sparsegs} attempt to directly use SDS to extract the external visual prior of the diffusion model, but have to suppress its weighting, thus achieving limited effects. Although view-conditioned diffusion priors \cite{zero123, zeronvs} have emerged recently, different to helpness for 3D generation \cite{zero123, invs}, their guidance is still limited for sparse-view reconstruction, which is detailedly discussed in the Appendix. Therefore, \textit{how to use diffusion priors} and \textit{how to use score distillation} under sparse views without fine-tuning, pre-training, and the optimization dilemma shown in Fig. \ref{fig:sds_trend} have become crucial issues.

\section{Method}
\label{sec:method}

With the phenomenon of the SDS dilemma shown in Fig. \ref{fig:sds_trend} in our mind, we have realized that SDS that works for text prompts does not work equally well for sparse views. Therefore, we attempt to analyze the disadvantages of SDS under sparse views and introduce inline constraints for effectively extracting visual guidance of diffusion priors without fine-tuning and pre-training. We start with the overview of 3DGS and also define the main symbols.

\subsection{Overview of 3D Gaussian Splatting}
\label{sec:3dgs}

\textbf{Representation}. 
The 3DGS models the 3D structure with a set of Gaussian points with positions $\mu_n$, covariance matrix $\Sigma_n$, color $c_n$ represented by Spherical Harmonic (SH) coefficients and opacity $\alpha_n$. For each Gaussian point $n$, its 3D position follows
\begin{equation}
    G(x) = e^{-\frac{1}{2}(x - \mu_n)^{\mathsf{T}} \Sigma_n^{-1} (x - \mu_n)},
\end{equation}
where $\Sigma_n$ can be represented by the scaling matrix $S_n$ and the rotation matrix $R_n$
\begin{equation}
    \Sigma_n = R_n S_n S_n^{\mathsf{T}} R_n^{\mathsf{T}}.
\end{equation}

\textbf{Rendering}.
For the 3D representation $\theta = \{\mu_n, \Sigma_n, c_n, \alpha_n\}$, we can optimize the trainable parameters $\theta$ through the following differentiable rendering function
\begin{equation}\label{eq:render_rgb}
    x_0(\mathbf{p}) = \sum_{n=1}^N c_n \tilde{\alpha}_n \prod_{m=1}^{n-1}(1-\tilde{\alpha}_m),
\end{equation}
where $x_0(\mathbf{p})$ is the rendering color at pixel $\mathbf{p}$ of rendered image $\mathbf{x}_0$, and $\tilde{\alpha}_n$ are computed from the projected 2D Gaussians.

\begin{figure}[t]
    \centering 
    \includegraphics[width=1.0\linewidth]{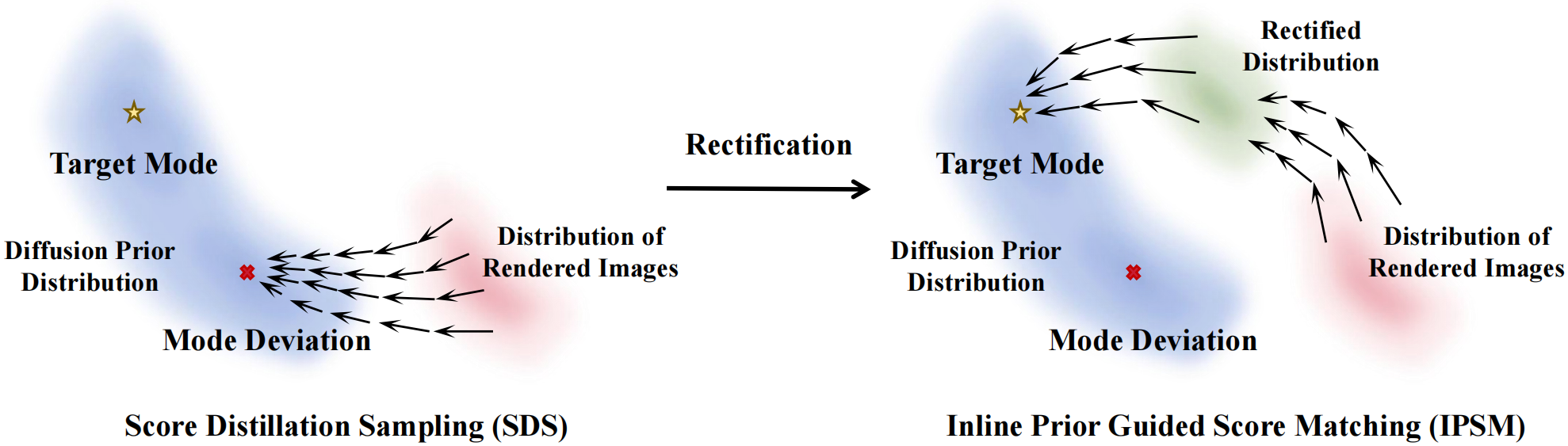} 
    \caption{\textbf{Comparison of SDS and IPSM.} \textbf{Left:} Tending to seek nearest mode, causing mode deviation. \textbf{Right:} Rectifying distribution to seek the target mode.} 
    \label{fig:rectify}
\end{figure}

\subsection{IPSM: Inline Prior Guided Score Matching}

\textbf{Review of Score Distillation Sampling}. 
Intuitively, SDS tends to drive the rendered image distribution denoted with \textcolor{render_distribution}{red color} seeking the nearest mode of diffusion distribution denoted with \textcolor{diffusion_prior}{blue color} guided by text prompts. 
Specifically, we denote the rendered image at viewpoint $\mathbf{v}^j$ as $\mathbf{x}_0^j = g(\theta, \mathbf{v}^j)$, where $g(\theta,\cdot)$ is rendering function and $\theta$ is the 3D representation needed optimization. Without elaborating text prompts on the conditions for brevity, the posterior noisy distribution of rendered images is defined as
\begin{equation}
    \textcolor{render_distribution}{q_t^{\theta}(\mathbf{x}_t^j)} \sim \mathcal{N}(\mathbf{x}_t^j; \sqrt{\Bar{\alpha}_t}\mathbf{x}_0^j, (1 - \Bar{\alpha}_t)\mathbf{I}).
\end{equation}
The prevailing score distillation works start from minimizing the reverse KL divergence between the distribution of the noisy rendered images $\textcolor{render_distribution}{q_t^{\theta}(\mathbf{x}_t^j)}$ and the noisy real-world distribution $\textcolor{diffusion_prior}{p_t^*(\mathbf{x}_t^j)}$ represented by the pre-trained diffusion models, namely
\begin{equation}\label{eq:sds_obj}
    \min_{\theta} \mathbb{E}_{t, \mathbf{v}_j}\bigg[ \omega(t) D_{KL}(\textcolor{render_distribution}{q_t^{\theta}(\mathbf{x}_t^j)} \Vert \textcolor{diffusion_prior}{p_t^*(\mathbf{x}_t^j)}) \bigg],
\end{equation}
which indicates the gradient of score distillation that
\begin{equation}\label{eq:sds_grad}
    \nabla_{\theta}\mathcal{L}_{\rm{SDS}}(\theta) \approx \mathbb{E}_{t, \boldsymbol{\epsilon}, \mathbf{v}^j} \bigg[ \omega(t) (\textcolor{diffusion_prior}{\boldsymbol{\epsilon}_{*}(\mathbf{x}_t^j, t)} - \textcolor{render_distribution}{\boldsymbol{\epsilon}}) \frac{\partial g(\theta, \mathbf{v}^j)}{\partial \theta} \bigg] = \mathbb{E}_{t, \boldsymbol{\epsilon}, \mathbf{v}^j} \bigg[ \frac{\omega(t)}{\gamma(t)} (\textcolor{render_distribution}{\mathbf{x}_0^j} - \textcolor{diffusion_prior}{\hat{\mathbf{x}}_0^{j;*}}) \frac{\partial g(\theta, \mathbf{v}^j)}{\partial \theta} \bigg],
\end{equation}
where $\gamma(t) = \frac{\sqrt{1 - \Bar{\alpha}_t}}{\sqrt{\Bar{\alpha}_t}}$ and $\textcolor{render_distribution}{\mathbf{x}_0^j \sim q_0^{\theta}(\mathbf{x}_0^j)}, \; \textcolor{diffusion_prior}{\hat{\mathbf{x}}_0^{j;*} \sim p_0^{*}(\mathbf{x}_0^j)}$. That is, for the given new viewpoints $\mathbf{v}_j$, the gradient $\nabla_{\theta}\mathcal{L}_{\rm{SDS}}(\theta)$ considers the rendered image distribution of the 3D representation and drives it closer to the pre-trained diffusion prior. 

\begin{figure}[t]
    \centering 
    \includegraphics[width=1.0\linewidth]{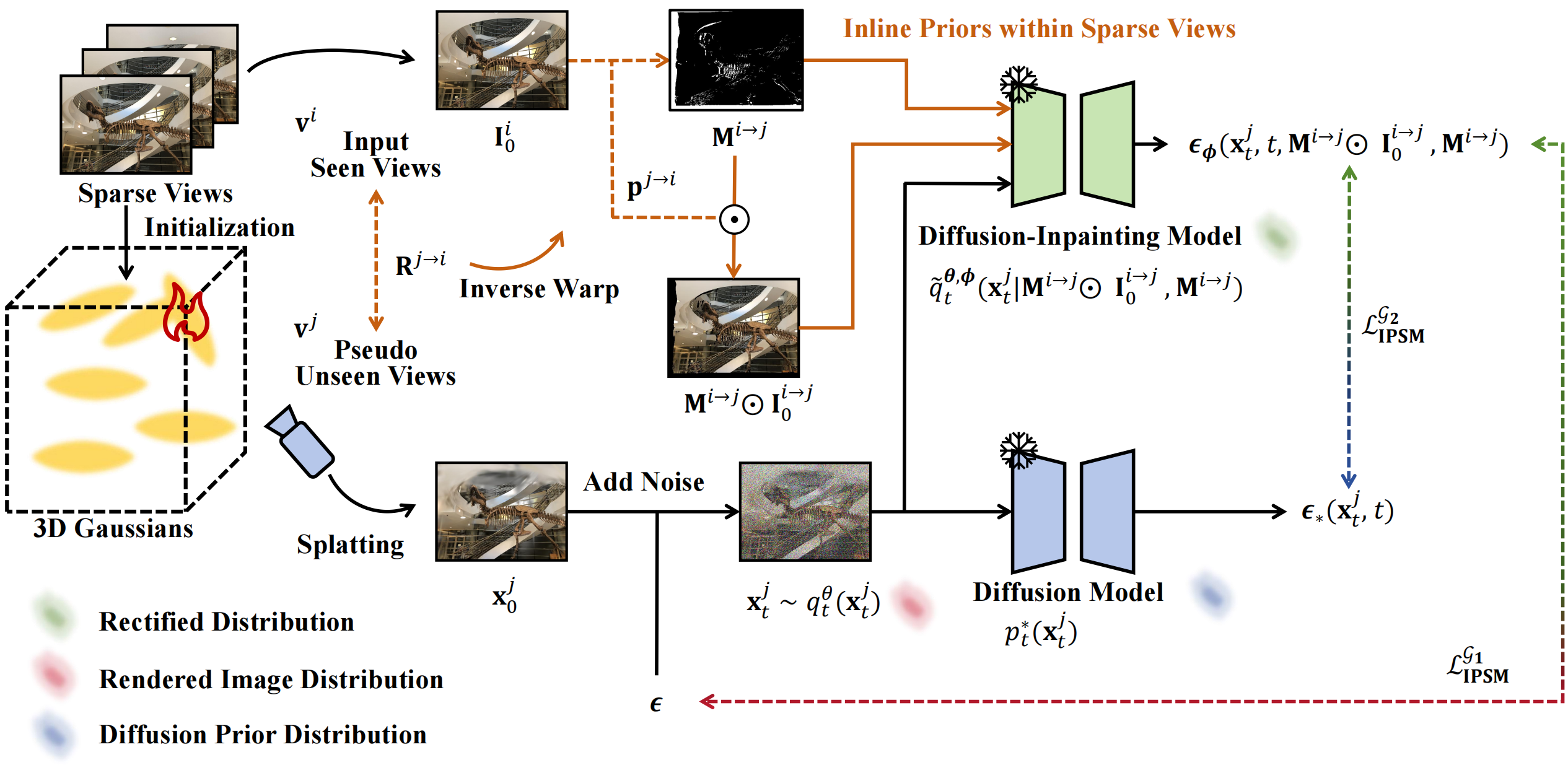} 
    \caption{\textbf{IPSM-Gaussian} obtains the \textcolor{inline_prior}{inline prior} within sparse views through inversely warping seen views to unseen pseudo views, thus modifying the \textcolor{render_distribution}{rendered image distribution} to the \textcolor{rectified_distribution}{rectified distribution}. Consequently taking the rectified distribution as the intermediate state, two sub-optimization objectives are utilized for controlling the optimization direction.} 
    \label{fig:pipeline}
\end{figure}

Following \cite{sds, stable_sds}, we provide a further discussion of SDS. The optimization objective of Eq. \ref{eq:sds_obj} derives $\textcolor{render_distribution}{q _ t^{\theta}(\mathbf{x} _ t^j)}$ to the high-density region of $\textcolor{diffusion_prior}{p _ t^*(\mathbf{x} _ t^j)}$. Considering samples $\textcolor{diffusion_prior}{\mathbf{m}^{\mathcal{T}}}, \textcolor{diffusion_prior}{\mathbf{m}^{\mathcal{F}}}$ from two modes of $\textcolor{diffusion_prior}{p _ t^*(\mathbf{x} _ t^j)}$, where $\textcolor{diffusion_prior}{\mathbf{m}^{\mathcal{T}}}$ is from target mode and $\textcolor{diffusion_prior}{\mathbf{m}^{\mathcal{F}}}$ is from failure mode.
$\textcolor{diffusion_prior}{\mathbf{m}^{\mathcal{F}}}$ is harmless for text-to-3D tasks due to the high information entropy properties of text prompts. However, for sparse-view 3D reconstruction, this leads the optimized 3D representation to be inconsistent with the given sparse images, thus causing optimization difficulties as shown in Fig. \ref{fig:rectify}. 
Specifically, we denote the L2 distance of two samples as $\Gamma(\cdot, \cdot)$. We want $\sqrt{\bar{\alpha} _ t} \textcolor{render_distribution}{\mathbf{x} _ 0 ^ j} \approx \sqrt{\bar{\alpha} _ t} \textcolor{diffusion_prior}{\mathbf{m}^{\mathcal{T}}}$ for any $t$, but the gap between two modes is unclear when $t$ increases, \ie $\Gamma( \sqrt{\bar{\alpha} _ t} \textcolor{render_distribution}{\mathbf{x} _ 0 ^ j}, \sqrt{\bar{\alpha} _ t} \textcolor{diffusion_prior}{\mathbf{m}^{\mathcal{T}}}) \approx \Gamma( \sqrt{\bar{\alpha} _ t} \textcolor{render_distribution}{\mathbf{x} _ 0 ^ j}, \sqrt{\bar{\alpha} _ t} \textcolor{diffusion_prior}{\mathbf{m}^{\mathcal{F}}} )$, since $| \Gamma( \textcolor{render_distribution}{\mathbf{x} _ 0 ^ j}, \textcolor{diffusion_prior}{\mathbf{m}^{\mathcal{T}}}) - \Gamma( \textcolor{render_distribution}{\mathbf{x} _ 0 ^ j}, \textcolor{diffusion_prior}{\mathbf{m}^{\mathcal{F}}} ) |$ is not large enough for a small $\sqrt{\bar{\alpha} _ t}$. This results in the mode aliasing for optimization and further affects the optimizing direction during training. To this end, the distribution of rendered images is not constrained to seeking the target mode, causing mode deviation. 
Therefore, we aim to construct a rectified distribution excluded failure mode using the inline prior from sparse views, whose sample $\textcolor{rectified_distribution}{\mathbf{m}^{\mathcal{R}}}$ provides $\textcolor{render_distribution}{\mathbf{x} _ 0 ^ j}$ stable optimization guidance and amplifies the gap $| \Gamma( \textcolor{rectified_distribution}{\mathbf{m}^{\mathcal{R}}}, \textcolor{diffusion_prior}{\mathbf{m}^{\mathcal{T}}}) - \Gamma( \textcolor{rectified_distribution}{\mathbf{m}^{\mathcal{R}}}, \textcolor{diffusion_prior}{\mathbf{m}^{\mathcal{F}}} ) |$ so that $\Gamma( \sqrt{\bar{\alpha}_t} \textcolor{rectified_distribution}{\mathbf{m}^{\mathcal{R}}}, \sqrt{\bar{\alpha}_t} \textcolor{diffusion_prior}{\mathbf{m}^{\mathcal{T}}} ) \ll \Gamma( \sqrt{\bar{\alpha}_t} \textcolor{rectified_distribution}{\mathbf{m}^{\mathcal{R}}}, \sqrt{\bar{\alpha}_t} \textcolor{diffusion_prior}{\mathbf{m}^{\mathcal{F}}} )$, and the rectified distribution is served as the bridge between $\textcolor{render_distribution}{\mathbf{x} _ 0 ^ j}$ and $\textcolor{diffusion_prior}{\mathbf{m}^{\mathcal{T}}}$ to control the mode-seeking direction.

\textbf{Inline Prior}. Different from text-to-3D tasks, sparse views can achieve geometry consistency guidance of novel views through camera pose transformation, namely the inline prior we mentioned in Sec. \ref{sec:intro}. Therefore, we aim to utilize the additional visual information of sparse views compared to text prompts to correct the erroneous tendency of SDS optimization. Specifically, we sample a set of random pseudo viewpoints $\mathbf{v}^j$ around the seen views $\mathbf{v}^i$. Given the ground-truth image $\mathbf{I}_0^i$ at the seen viewpoint $\mathbf{v}^i$, we formulate the transforming function $\psi(\mathbf{I}_0^i; \mathbf{D}^j, \mathbf{R}^{j \rightarrow i})$ which inversely warps image $\mathbf{I}_0^i$ from viewpoint $\mathbf{v}^i$ to $\mathbf{v}^j$. $\mathbf{R}^{j \rightarrow i}$ represents the relative pose transformation between two viewpoints, and $\mathbf{D}^j$ is the alpha-blending rendered depth at viewpoint $\mathbf{v}^j$ following
\begin{equation}\label{eq:render_depth}
    D^j(\mathbf{p}) = \sum_{n=1}^{N} d_n \tilde{\alpha}_n \prod_{m=1}^{N-1}(1 - \tilde{\alpha}_{m}), 
\end{equation}
where $d_n$ is the z-buffer of the $n$-th Gaussian. During transformation, each pixel location $\mathbf{p}^j$ at the pseudo viewpoint $\mathbf{v}^j$ is warped to the pixel location $\mathbf{p}^{j \rightarrow i}$ at the seen viewpoint $\mathbf{v}^i$, and $\mathbf{p}^{j \rightarrow i}$ can be represented by
\begin{equation}
\label{eq:warp_pji}
    \mathbf{p}^{j \rightarrow i} \sim \mathbf{K} \mathbf{R}^{j \rightarrow i} D^j(\mathbf{p}^j) \mathbf{K}^{-1} \mathbf{p}^j, 
\end{equation}
where $\mathbf{K}$ is the camera intrinsic parameter. Then, we can obtain the warped image $I_0^{i \rightarrow j}(\mathbf{p}^j)$ using inverse warping with the nearest sampling operator
\begin{equation}\label{eq:inverse_warp}
    I_0^{i \rightarrow j}(\mathbf{p}^j) = {\rm{Sampler}}(\mathbf{I}_0^i, \mathbf{p}^{j \rightarrow i}).
\end{equation}
However, this direct inverse warping may lead to warping distortion due to erroneous geometry. Following \cite{geconerf}, we tackle it through the generated consistency mask with an error threshold $\tau$
\begin{equation}\label{eq:warp_mask}
    M^{i \rightarrow j}(\mathbf{p}^j) = {\rm{Mask}}(\Vert D^j(\mathbf{p}^j) - D^{i \rightarrow j}(\mathbf{p}^j) \Vert_1 < \tau),
\end{equation}
where $D^{i \rightarrow j}(\mathbf{p}^j) = {\rm{Sampler}}(\mathbf{D}^i, \mathbf{p}^{j \rightarrow i})$ like Eq. \ref{eq:inverse_warp}. Eq. \ref{eq:warp_mask} ensures the filterability of erroneous geometry using the difference between the warped depth of the seen viewpoint and the depth of the pseudo viewpoint. In practice, the warped image $\mathbf{I}_0^{i \rightarrow j}$ and its accompanying mask $\mathbf{M}^{i \rightarrow j}$ are served as the \textit{inline geometry consistency prior} to guide external diffusion prior scene specialization. The intuitive explanation of inline priors can be found in Appendix \ref{sec:appx:vis_inline}.

\textbf{Inline Prior Guided Score Matching}. Using score distillation directly in the case of sparse views overlooks the inline geometry consistency prior within the sparse views themselves, which is fundamentally different from text-to-3D. 
To this end, we rectify the distribution denoted with \textcolor{rectified_distribution}{green color} from $\textcolor{render_distribution}{q_0^{\theta}(\mathbf{x}_0^j)}$ to $\textcolor{rectified_distribution}{\tilde{q}_0^{\theta, \phi}(\mathbf{x}_0^j | \mathbf{M}^{i \rightarrow j} \odot \mathbf{I}_0^{i \rightarrow j}, \mathbf{M}^{i \rightarrow j})}$ using the inline prior.
As shown in Fig. \ref{fig:pipeline}, we utilize the warped masked image $\mathbf{I}_0^{i \rightarrow j}$ from the seen viewpoints to guide the sampling trajectory of $\textcolor{rectified_distribution}{\hat{\mathbf{x}}_0^{j;\phi} \sim \tilde{q}_0^{\theta, \phi}(\mathbf{x}_0^j | \mathbf{M}^{i \rightarrow j} \odot \mathbf{I}_0^{i \rightarrow j}, \mathbf{M}^{i \rightarrow j})}$, thus introducing the inline geometry consistency prior to the score distillation. 
So our optimization objective is changed to minimizing (1) the KL divergence between the noisy rendered image distribution $\textcolor{render_distribution}{q_t^{\theta}(\mathbf{x}_t^j)}$ and the noisy rectified distribution $\textcolor{rectified_distribution}{\tilde{q}_t^{\theta, \phi}(\mathbf{x}_t^j)}$; (2) the KL divergence between the noisy rectified distribution $\textcolor{rectified_distribution}{\tilde{q}_t^{\theta, \phi}(\mathbf{x}_t^j)}$ and the noisy diffusion prior distribution $\textcolor{diffusion_prior}{p_t^*(\mathbf{x}_t^j)}$ represented by the pre-trained diffusion models, namely
\begin{equation}
    \min_{\theta} \left\{ \eta_r \mathbb{E}_{t, c}\bigg[ \omega(t) D_{KL}(\textcolor{render_distribution}{q_t^{\theta}(\mathbf{x}_t^j)} \Vert \textcolor{rectified_distribution}{\tilde{q}_t^{\theta, \phi}(\mathbf{x}_t^j))} \bigg] + \mathbb{E}_{t, c}\bigg[ \omega(t) D_{KL}(\textcolor{rectified_distribution}{\tilde{q}_t^{\theta, \phi}(\mathbf{x}_t^j)} \Vert \textcolor{diffusion_prior}{p_0^*(\mathbf{x}_t^j)}) \bigg] \right\},
\end{equation}
where $\eta_r$ is the adjustment parameter of the two sub-optimization objectives.
In practice, we introduce an inpainting diffusion model $\textcolor{rectified_distribution}{\boldsymbol{\epsilon}_{\phi}(\mathbf{x}_t^j, t, \mathbf{M}^{i \rightarrow j} \odot \mathbf{I}_0^{i \rightarrow j}, \mathbf{M}^{i \rightarrow j})}$, which shares the same VAE-feature domain with the pre-trained diffusion model $\textcolor{diffusion_prior}{\boldsymbol{\epsilon}_*(\mathbf{x}_t^j, t)}$ representing the real data distribution. So we have the rectified gradient of score distillation
\begin{equation}
    \begin{aligned}
        \nabla_{\theta}\mathcal{L}_{\rm{IPSM}}(\theta) \approx &\eta_r \mathbb{E}_{t, \boldsymbol{\epsilon}, \mathbf{v}^j} \bigg[ \frac{\omega(t)}{\gamma(t)} (\textcolor{render_distribution}{\mathbf{x}_0^j} - \textcolor{rectified_distribution}{\hat{\mathbf{x}}_0^{j;\phi}}) \frac{\partial g(\theta, \mathbf{v}^j)}{\partial \theta} \bigg] + \mathbb{E}_{t, \boldsymbol{\epsilon}, \mathbf{v}^j} \bigg[ \frac{\omega(t)}{\gamma(t)} (\textcolor{rectified_distribution}{\hat{\mathbf{x}}_0^{j;\phi}} - \textcolor{diffusion_prior}{\hat{\mathbf{x}}_0^{j;*})} \frac{\partial g(\theta, \mathbf{v}^j)}{\partial \theta} \bigg] \\
        = &\eta_r \mathbb{E}_{t, \boldsymbol{\epsilon}, \mathbf{v}^j} \bigg[ \omega(t) (\textcolor{rectified_distribution}{\boldsymbol{\epsilon}_{\phi}(\mathbf{x}_t^j, t, \mathbf{M}^{i \rightarrow j} \odot \mathbf{I}_0^{i \rightarrow j}, \mathbf{M}^{i \rightarrow j})} - \textcolor{render_distribution}{\boldsymbol{\epsilon}}) \frac{\partial g(\theta, \mathbf{v}_j)}{\partial \theta} \bigg] \\
        &+ \mathbb{E}_{t, \boldsymbol{\epsilon}, \mathbf{v}^j} \bigg[ \omega(t) (\textcolor{diffusion_prior}{\boldsymbol{\epsilon}_{*}(\mathbf{x}_t^j, t)} - \textcolor{rectified_distribution}{\boldsymbol{\epsilon}_{\phi}(\mathbf{x}_t^j, t, \mathbf{M}^{i \rightarrow j} \odot \mathbf{I}_0^{i \rightarrow j}, \mathbf{M}^{i \rightarrow j})}) \frac{\partial g(\theta, \mathbf{v}^j)}{\partial \theta} \bigg].
    \end{aligned}
\end{equation}
Consequently, the IPSM regularization can be represented as
\begin{equation}\label{eq:loss_ipsm}
    \mathcal{L}_{\rm{IPSM}} = \eta_r \underbrace{\mathbb{E}_{t, \boldsymbol{\epsilon}, \mathbf{v}^j} \bigg[ \Vert \omega(t) (\textcolor{rectified_distribution}{\boldsymbol{\epsilon}_{\phi}} - \textcolor{render_distribution}{\boldsymbol{\epsilon}}) \Vert_2^2 \bigg]}_{\mathcal{L}_{\rm{IPSM}}^{\mathcal{G}_1}}+ \underbrace{\mathbb{E}_{t, \boldsymbol{\epsilon}, \mathbf{v}^j} \bigg[ \Vert \omega(t) (\textcolor{diffusion_prior}{\boldsymbol{\epsilon}_{*}} - \textcolor{rectified_distribution}{\boldsymbol{\epsilon}_{\phi}}) \Vert_2^2 \bigg]}_{\mathcal{L}_{\rm{IPSM}}^{\mathcal{G}_2}}.
\end{equation}

\subsection{Training Details}
\label{sec:training_details}

\textbf{Depth Regularization}. In the warping process, it can be observed that the rendered depth influences pixel mapping relations, which is detailed in Sec. \ref{sec:exp}. Therefore, it is necessary to incorporate monocular depth estimation prior to supervising rendered depth, thus providing the correct inline prior. We use the Pearson Correlation to provide depth regularization, which can be represented as
\begin{equation}
    {{\rm{Corr}}(\mathbf{D}_{r}, \mathbf{D}_{m})} = \frac{{\rm{Cov}}(\mathbf{D}_r, \mathbf{D}_m)}{\sqrt{{\rm{Var}}(\mathbf{D}_r) {\rm{Var}}(\mathbf{D}_m)}}.
\end{equation}
Given the rendered depth $\mathbf{D}_r^i$, monocular depth $\mathbf{D}_m^i$ from the input view $\mathbf{I}_0^i$ at the seen view $\mathbf{v}^i$, and the rendered depth $\mathbf{D}_r^j$, monocular depth $\mathbf{D}_m^j$ from the rendered image $\mathbf{x}_0^j$ at the unseen view $\mathbf{v}^j$, we take the depth regularization as
\begin{equation}
    \mathcal{L}_{\rm{depth}} = \eta_d \Vert {{\rm{Corr}}(\mathbf{D}_r^i, \mathbf{D}_m^i)} \Vert_1 + \Vert {{\rm{Corr}}(\mathbf{D}_r^j, \mathbf{D}_m^j)} \Vert_1,
\end{equation}
where $\eta_d$ serves as the weight to balance the supervision of seen views and pseudo-unseen views.

\textbf{Geometry Consistency Regularization}. In Eq. \ref{eq:loss_ipsm}, we introduce $\mathcal{L}_{\rm{IPSM}}^{\mathcal{G}_1}$ for providing guidance to minimize the reverse KL divergence between the rendered image and rectified distribution. In practice, we not only supervise from the diffusion feature domain but also provide stronger guidance by directly adding masked L1 loss of $\mathbf{x}_0^j$ and $\mathbf{I}_0^{i \rightarrow j}$, which is denoted as the geometry consistency regularization and can be represented as
\begin{equation}
    \mathcal{L}_{\rm{geo}} = \Vert \mathbf{M}^{i \rightarrow j} \odot (\mathbf{x}_0^j - \mathbf{I}_0^{i \rightarrow j}) \Vert_1.
\end{equation}

\textbf{Total Training Objectives}.
Overall, our training objectives can be divided into three parts: 1) The direct supervision $\mathcal{L}_1$ and $\mathcal{L}_{\rm{ssim}}$ of the sparse input views, which are inherited from the vanilla 3DGS; 2) The supervision $\mathcal{L}_{\rm{IPSM}}$ provided by diffusion priors using IPSM; 3) The supervision of depth and vision information $\mathcal{L}_{\rm{depth}}$ and $\mathcal{L}_{\rm{geo}}$ to support the inline priors and provide low-level inline guidance. The total training loss function can be summarized as
\begin{equation}
    \mathcal{L} = \lambda_1 \mathcal{L}_1 + \lambda_{\rm{ssim}} \mathcal{L}_{\rm{ssim}} + \lambda_{\rm{depth}} \mathcal{L}_{\rm{depth}} + \lambda_{\rm{geo}} \mathcal{L}_{\rm{geo}} + \lambda_{\rm{IPSM}} \mathcal{L}_{\rm{IPSM}}.
\end{equation}
More training details are shown in the Appendix \ref{sec:appx:training_details}.

\begin{table}[t]
\centering
\caption{\textbf{Quantitative comparisons with other methods.}}\label{tab:comp_sota}
\setlength{\tabcolsep}{0.6mm}{
\begin{tabular}{c|c|cccc|cccc}
\toprule
\multirow{2}{*}{Methods} & \multirow{2}{*}{Setting} & \multicolumn{4}{c|}{LLFF \cite{llff}} & \multicolumn{4}{c}{DTU \cite{dtu}} \\ \cmidrule{3-10} 
& & \multicolumn{1}{l}{SSIM\textuparrow} & \multicolumn{1}{l}{LPIPS\textdownarrow} & \multicolumn{1}{l}{PSNR\textuparrow} & \multicolumn{1}{l|}{AVGE\textdownarrow} & \multicolumn{1}{l}{SSIM\textuparrow} & \multicolumn{1}{l}{LPIPS\textdownarrow} & \multicolumn{1}{l}{PSNR\textuparrow} & \multicolumn{1}{l}{AVGE\textdownarrow} \\ \midrule
SRF \cite{srf} & \multirow{3}{*}{\begin{tabular}[c]{@{}c@{}}Trained on \\DTU\end{tabular}} & 0.250 & 0.591 & 12.34 & 0.313 & 0.671 & 0.304 & 15.32 & 0.171 \\
PixelNeRF \cite{pixelnerf} & & 0.272 & 0.682 & 7.93 & 0.461 & 0.695 & 0.270 & 16.82 & 0.147 \\
MVSNeRF \cite{mvsnerf} & & 0.557 & 0.356 & 17.25 & 0.171 & 0.769 & 0.197 & 18.63 & 0.113 \\ \hline
SRF ft. \cite{srf} & \multirow{3}{*}{\begin{tabular}[c]{@{}c@{}}Fine-tuned\\ per Scene\end{tabular}} & 0.436 & 0.529 & 17.07 & 0.203 & 0.698 & 0.281 & 15.68 & 0.162 \\
PixelNeRF ft. \cite{pixelnerf} & & 0.438 & 0.512 & 16.17 & 0.217 & 0.710 & 0.269 & 18.95 & 0.125 \\
MVSNeRF ft. \cite{mvsnerf} & & 0.584 & 0.327 & 17.88 & 0.157 & 0.769 & 0.197 & 18.54 & 0.113 \\ \hline
Mip-NeRF \cite{mipnerf} & \multirow{5}{*}{\begin{tabular}[c]{@{}c@{}}Based on\\ NeRF\\ Optimized\\ per Scene\end{tabular}} & 0.351 & 0.495 & 14.62 & 0.246 & 0.571 & 0.353 & 8.68 & 0.323 \\
DietNeRF \cite{dietnerf} & & 0.370 & 0.496 & 14.94 & 0.240 & 0.633 & 0.314 & 11.85 & 0.243 \\
RegNeRF \cite{regnerf} & & 0.587 & 0.336 & 19.08 & 0.149 & 0.745 & 0.190 & 18.89 & 0.112 \\
FreeNeRF \cite{freenerf} & & 0.612 & 0.308 & 19.63 & 0.134 & 0.787 & 0.182 & \cellcolor{second}19.92 & \cellcolor{second}0.098 \\
SparseNeRF \cite{sparsenerf} & & 0.624 & 0.328 & \cellcolor{third}19.86 & 0.127 & 0.769 & 0.201 & \cellcolor{third}19.55 & 0.102 \\ \hline
3DGS \cite{3dgs} & \multirow{6}{*}{\begin{tabular}[c]{@{}c@{}}Based on\\ 3DGS\\ Optimized\\ per Scene\end{tabular}} & 0.456 & 0.385 & 14.97 & 0.208 & \cellcolor{third}0.795 & 0.178 & 15.06 & 0.136 \\
FSGS \cite{fsgs} & & \cellcolor{third}0.682 & \cellcolor{third}0.248 & \cellcolor{second}20.43 & \cellcolor{second}0.108 & \cellcolor{second}0.825 & \cellcolor{second}0.145 & 17.69 & \cellcolor{third}0.101 \\
DNGaussian \cite{dngaussian} & & 0.591 & 0.294 & 19.12 & 0.132 & 0.790 & \cellcolor{third}0.176 & 18.91 & 0.102 \\
DNGaussian \dag \cite{dngaussian} & & \cellcolor{second}0.687 & \cellcolor{second}0.228 & 19.94 & \cellcolor{third}0.109 & - & - & - & - \\
\multirow{2}{*}{Ours} & & \cellcolor{first}0.702 & \cellcolor{first}0.207 & \cellcolor{first}20.44 & \cellcolor{first}0.101 & \cellcolor{first}0.856 & \cellcolor{first}0.121 & \cellcolor{first}19.99 & \cellcolor{first}0.077 \\ 
& & \cellcolor{first}\textpm 0.001 & \cellcolor{first}\textpm 0.001 & \cellcolor{first}\textpm 0.08 & \cellcolor{first}\textpm 0.001 & \cellcolor{first}\textpm 0.001 & \cellcolor{first}\textpm 0.001 & \cellcolor{first}\textpm 0.10 & \cellcolor{first}\textpm 0.001 \\ \bottomrule
\end{tabular}
}
\begin{tablenotes}
    \item[\dag] \dag: Using SfM initialization same as 3DGS, FSGS and Ours for fair comparisons.
\end{tablenotes}
\end{table}

\section{Experiments}\label{sec:exp}

\subsection{Experiments Settings}\label{sec:exp_settings}

\textbf{Datasets and Metrics}.
We evaluate our method on the LLFF \cite{llff} and DTU dataset \cite{dtu}. The LLFF dataset involves 8 forward-facing scenes and we select 3 training views following prevailing works \cite{freenerf, regnerf}. On the DTU dataset, we choose the 15 testing scenes, and 3 training views whose IDs are 25, 22, and 28, following RegNeRF \cite{regnerf}. Following prevailing works \cite{regnerf, freenerf, dngaussian} to focus on the object-of-interest for the DTU dataset, we also remove the background with the mask of objects when evaluating. Aligning with the protocol of baselines, we apply the downsampling rate of 8 and 4 on the LLFF and DTU datasets respectively.
We evaluate the reconstruction quality using SSIM \cite{ssim}, LPIPS \cite{lpips}, and PSNR. Following DNGaussian \cite{dngaussian} and FreeNeRF \cite{freenerf}, we also report AVGE for a comprehensive evaluation of the reconstruction quality. The AVGE is calculated by the geometric mean of $\sqrt{1 - {\text{SSIM}}}$, $\text{LPIPS}$, and ${\text{MSE}}=10^{-{\text{PSNR}}/10}$. The experiments are conducted 3 times and we report the mean and standard deviation. More details about datasets, \eg the sparsity of training views and train-test split protocols, can be found in Appendix \ref{sec:appx:dataset_details}.

\textbf{Implementation details}. 
Our method is built on 3DGS instead of NeRF due to the advantages of 3DGS on high training speed and real-time rendering. Following prevailing works \cite{freenerf, dngaussian}, the camera poses are known before optimization. The initialized point clouds are estimated by Structure from Motion (SfM) \cite{sfm} only using the given sparse input views. The total training process involves 10K iterations for experiments on all datasets. The guidance of pseudo views starts from 2K iteration and ends at 9.5K iteration. Following FSGS, we introduce the proximity-guided Gaussian unpooling operation \cite{fsgs} and retain the high tolerance for large Gaussian points without size thresholds. For the score distillation methods, we randomly select one of 3 training views to generate BLIP-based \cite{blip} text prompts. Background priors are introduced on DTU for accurately reconstructing the object-of-interest. All experimental results are obtained on a single RTX 3090. More training details and experimental environments can be found in Appendix \ref{sec:appx:training_details} and \ref{sec:appx:hyperparameters}.

\textbf{Baselines}. 
Following prevailing works, we compare our method with the state-of-the-art methods, \ie SRF \cite{srf}, PixelNeRF \cite{pixelnerf}, MVSNeRF \cite{mvsnerf}, Mip-NeRF \cite{mipnerf}, DietNeRF \cite{dietnerf}, RegNeRF \cite{regnerf}, FreeNeRF \cite{freenerf}, SparseNeRF \cite{sparsenerf}, the vanilla 3DGS \cite{3dgs}, FSGS \cite{fsgs} and DNGaussian \cite{dngaussian} as our baselines. 
Except for the reproduced results of the 3DGS \cite{3dgs} on the LLFF dataset, and 3DGS \cite{3dgs} and FSGS \cite{fsgs} on the DTU dataset, the rest are based on the values reported. Since the original DNGaussian uses random initialization, while other 3DGS methods use SfM \cite{sfm}, we also report the provided LLFF results of using SfM \cite{sfm}. Reproduction details can be found in the Appendix \ref{sec:appx:baselines}.

\subsection{Comparison with Other Methods}
\label{sec:comparison_with_others}

\begin{figure}[t]
    \centering 
    \includegraphics[width=1.0\linewidth]{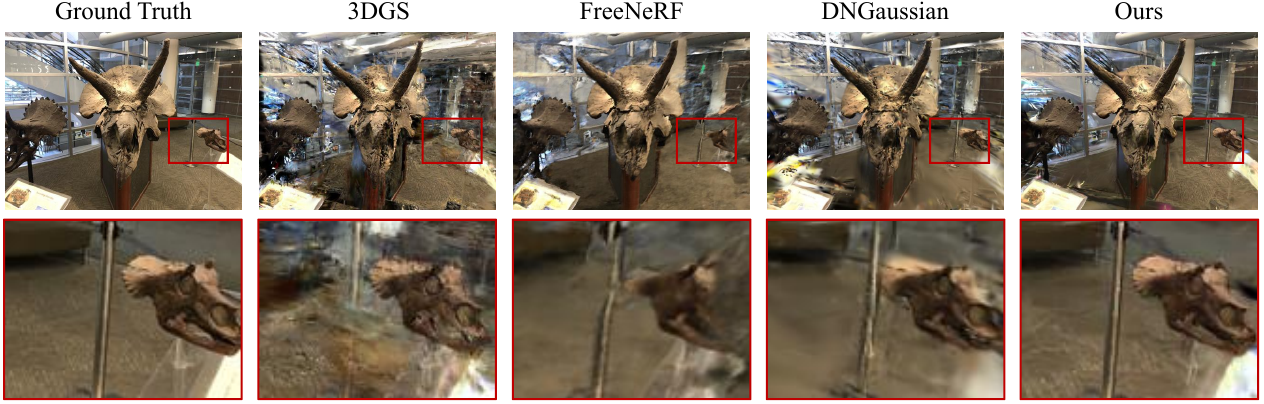} 
    \caption{\textbf{Qualitative comparison on the LLFF dataset}.} 
    \label{fig:comp_sota_llff}
\end{figure}

\textbf{LLFF}. 
The quantitative results on the LLFF dataset \cite{llff} are shown in Tab. \ref{tab:comp_sota}. Our method shows significant improvement and achieves the best reconstruction quality among state-of-the-art methods under multi-metric evaluation. For the NeRF-based methods, SSIM of our method is improved by $+12.5\%$ compared to SparseNeRF \cite{sparsenerf}, and LPIPS is improved by $+32.79\%$ compared to FreeNeRF \cite{freenerf}, which are the state-of-the-art in the NeRF-based methods respectively. For the 3DGS-based methods, the AVGE of our method is improved by $+6.48\%$ and $+7.34\%$ compared to the state-of-the-art FSGS \cite{fsgs} and DNGaussian \dag \cite{dngaussian} respectively. Note that the vanilla DNGaussian uses random initialization, but the 3DGS, FSGS, and our method use SfM initialization. Thus, we also report the provided results of SfM-initialized DNGaussian which is denoted by \dag. The qualitative results are shown in Fig. \ref{fig:comp_sota_llff}. Due to the lack of external priors, 3DGS \cite{3dgs} and FreeNeRF \cite{freenerf} show the optimization tendencies of 3D representations themselves, which are high-frequency artifacts and low-frequency smoothness respectively. Although DNGaussian \cite{dngaussian} using external depth prior can suppress artifacts, it only uses coarse-grained depth guidance and lacks fine-grained visual guidance, so the rendered image lacks high-frequency information. Our approach achieves improvements in both visual and geometric quality.

\begin{wrapfigure}{r}{7cm}
    \centering 
    \vspace{-1.5em}
    \includegraphics[width=1.0\linewidth]{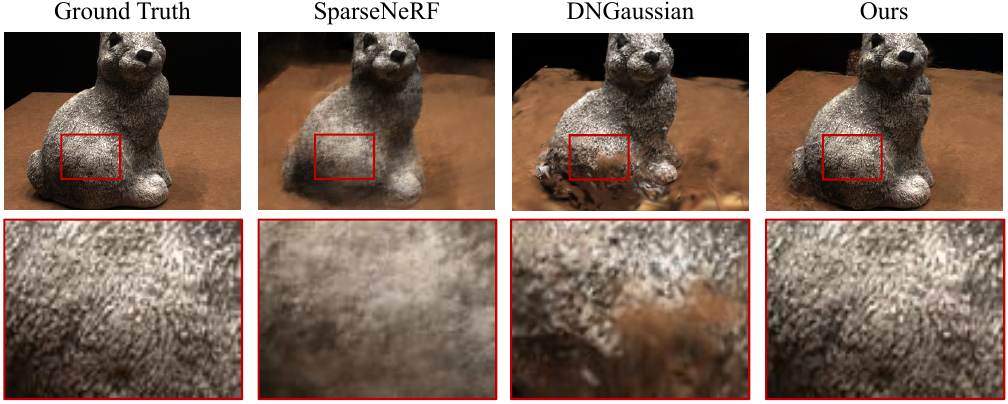} 
    \caption{\textbf{Qualitative comparison on DTU}.} 
    \label{fig:comp_sota_dtu}
    \vspace{-1.5em}
\end{wrapfigure}

\textbf{DTU}. Similar performances of the quantitative results on the DTU dataset \cite{dtu} are shown in Tab. \ref{tab:comp_sota}.
The AVGE of our method is improved by $+23.76\%$ compared to FSGS \cite{fsgs} and $+21.43\%$ compared to FreeNeRF \cite{freenerf}. 
Note that DNGaussian \cite{dngaussian} does not provide the corresponding parameter settings for using SfM \cite{sfm} initialization on the DTU dataset \cite{dtu}. The qualitative results are shown in Fig. \ref{fig:comp_sota_dtu}. SparseNeRF \cite{sparsenerf} and DNGaussian \cite{dngaussian}, which only use depth priors, cannot obtain guidance on visual texture details, causing optimization difficulties. Our IPSM-Gaussian using diffusion priors can obtain textured details of reconstruction close to the Ground Truth.

Details of reported experimental results are shown in Appendix \ref{sec:appx:details_exp_results}. More rendered novel views and qualitative comparisons can be found in the Appendix \ref{sec:appx:qualitative_results}.

\subsection{Ablation Study}\label{sec:ablation}

\begin{table}[t]
\centering
\caption{\textbf{Ablation Study} on the LLFF dataset with 3-views setting.}
\label{tab:comp_abla}
\setlength{\tabcolsep}{0.8mm}{
\begin{tabular}{cccccccc}
\toprule
\multicolumn{2}{c}{w/ $\mathcal{L}_{\rm{IPSM}}$} & \multirow{2}{*}{w/ $\mathcal{L}_{\rm{depth}}$} & \multirow{2}{*}{w/ $\mathcal{L}_{\rm{geo}}$} & \multirow{2}{*}{SSIM\textuparrow} & \multirow{2}{*}{LPIPS\textdownarrow} & \multirow{2}{*}{PSNR\textuparrow} & \multirow{2}{*}{AVGE\textdownarrow} \\ 
w/ $\mathcal{L}_{\rm{IPSM}}^{\mathcal{G}_1}$ & w/ $\mathcal{L}_{\rm{IPSM}}^{\mathcal{G}_2}$ & & & & &  \\ \midrule
& & & & 0.625 \textpm 0.008 & 0.254 \textpm 0.007 & 19.00 \textpm 0.12 & 0.125 \textpm 0.003 \\ 
\checkmark & & & & 0.636 \textpm 0.004 & 0.245 \textpm 0.003 & 19.22 \textpm 0.02 & 0.121 \textpm 0.001 \\ 
\checkmark & \checkmark & & & 0.670 \textpm 0.001 & 0.229 \textpm 0.002 & 19.60 \textpm 0.11 & 0.113 \textpm 0.001 \\
\checkmark & \checkmark & \checkmark & & 0.697 \textpm 0.002 & 0.211 \textpm 0.001 & 20.20 \textpm 0.03 & 0.104 \textpm 0.001 \\ 
\checkmark & \checkmark & \checkmark & \checkmark & 0.702 \textpm 0.001 & 0.207 \textpm 0.001 & 20.44 \textpm 0.08 & 0.101 \textpm 0.001 \\ \bottomrule
\end{tabular}
}
\end{table}

\begin{figure}[t]
    \centering 
    \includegraphics[width=1.0\linewidth]{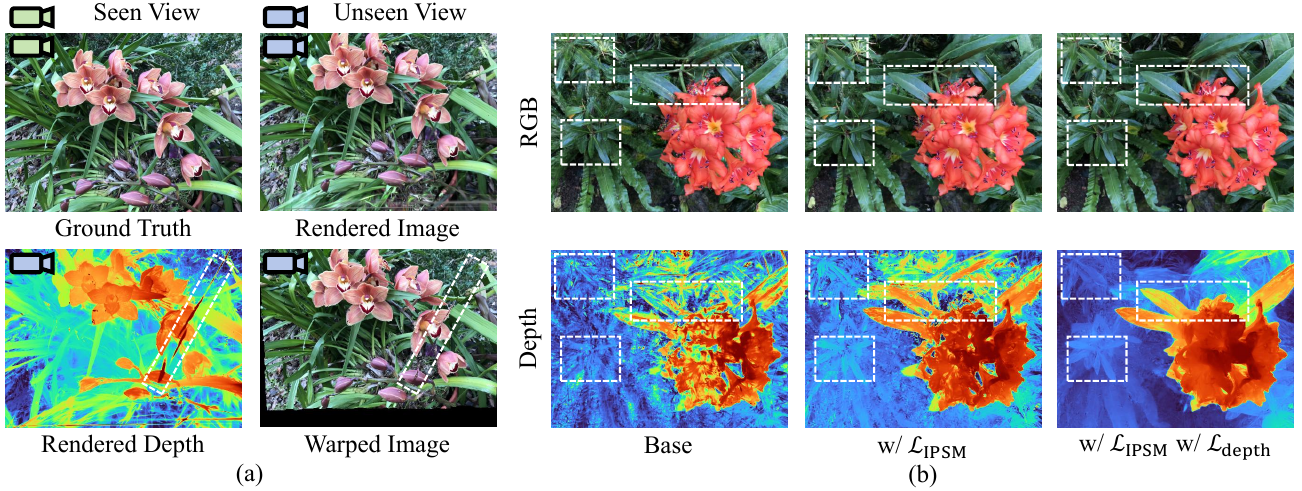} 
    \caption{\textbf{(a)} Impact of depth error on the inline prior. \textbf{(b)} Ablation of IPSM and depth regularizations.} 
    \label{fig:comp_abla}
\end{figure}

We conduct detailed ablations of regularization terms on the LLFF dataset \cite{llff} shown in Tab. \ref{tab:comp_abla}. 
We can notice that the first two regularization terms, \ie IPSM and depth, provide significant improvements.
The first three lines demonstrate the promoting effect of our proposed IPSM on the reconstruction quality of 3D representations, \eg using IPSM boosts $9.8\%$ on the LPIPS and $9.6\%$ on the AVGE compared to the Base. It is worth noting that since the inline prior requires an accurate rendering depth from the unseen perspective shown in Eq. \ref{eq:warp_pji}. The impact of depth error on inline priors is shown in Fig. \ref{fig:comp_abla} (a). However, the diffusion priors, as a kind of visual supervision, cannot provide direct depth geometry guidance, so an additional external depth prior needs to be introduced, which can support the accuracy of inline prior to further provide performance improvements. In Fig. \ref{fig:comp_abla} (b), we show the visual and geometry improvements of IPSM and depth regularization. The last line in Tab. \ref{tab:comp_abla} introduces the geometry consistency regularization for providing pixel-wise guidance, which shows a steady improvement. More additional ablations are detailed in the Appendix \ref{sec:appx:additional_ablation}.

\subsection{Comparison to SDS}\label{sec:comp_sds}

\begin{figure}[t]
    \centering 
    \includegraphics[width=1.0\linewidth]{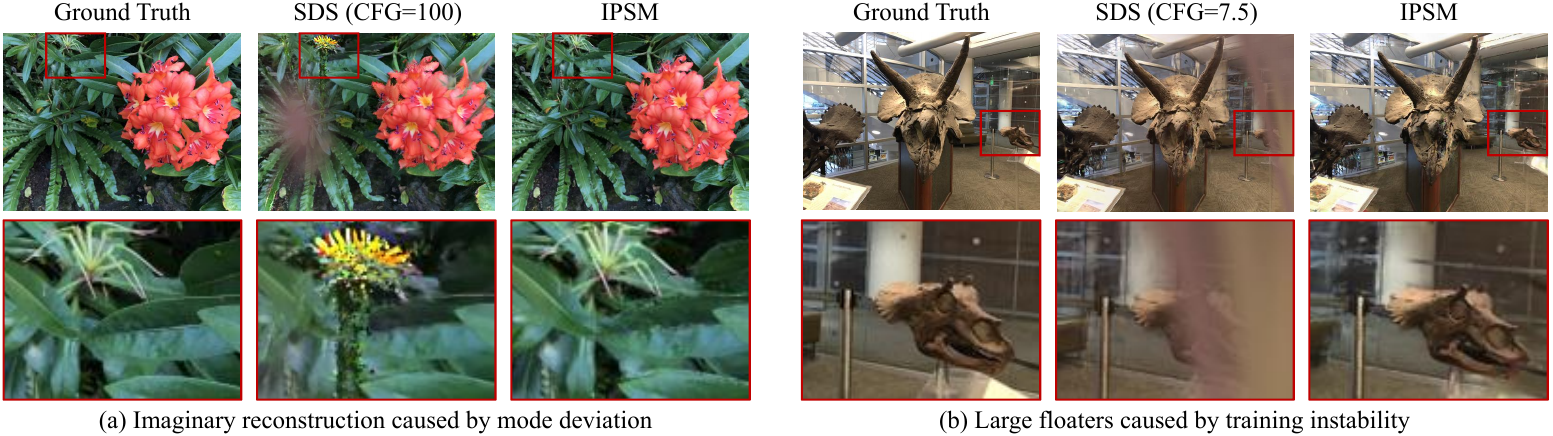}
    \caption{Qualitative comparison with SDS.} 
    \label{fig:qualitative_sds}
\end{figure}
\begin{table}[t]
\centering
\caption{\textbf{Comparison to SDS} on the LLFF dataset with 3-views setting.}
\label{tab:comp_sds}
\setlength{\tabcolsep}{1.9mm}{
\begin{tabular}{ccccc}
\toprule
Setting & SSIM\textuparrow & LPIPS\textdownarrow & PSNR\textuparrow & AVGE\textdownarrow \\ \midrule
Base & 0.625 \textcolor{gray}{(+0.00\%)} & 0.254 \textcolor{gray}{(+0.00\%)} & 19.00 \textcolor{gray}{(+0.00\%)} & 0.125 \textcolor{gray}{(+0.00\%)} \\ 
w/ SDS(CFG=7.5) & 0.647 \textcolor{red}{(+3.52\%)} & 0.267 \textcolor{blue}{(-5.12\%)} & 18.80 \textcolor{blue}{(-1.05\%)} & 0.128 \textcolor{blue}{(-2.40\%)} \\ 
w/ SDS(CFG=100) & 0.576 \textcolor{blue}{(-7.84\%)} & 0.367 \textcolor{blue}{(-44.49\%)} & 17.53 \textcolor{blue}{(-7.74\%)} & 0.162 \textcolor{blue}{(-29.60\%)} \\ 
w/ IPSM(CFG=7.5) & \textbf{0.670 \textcolor{red}{(+7.20\%)}} & \textbf{0.229 \textcolor{red}{(+9.84\%)}} & \textbf{19.60 \textcolor{red}{(+3.16\%)}} & \textbf{0.113 \textcolor{red}{(+9.60\%)}} \\ \bottomrule
\end{tabular}
}
\end{table}

As shown in Fig. \ref{fig:sds_trend}, SDS guidance is hard to provide effective supervision but tends to hinder reconstruction due to the mode deviation we have analyzed. Due to the too-strong semantic visual supervision of SDS(CFG=100), the performance increases significantly in the final 500 iterations after the 2K-9.5K prior-added period instead. In this section, we report the final evaluated performance comparison of \textit{Base} (without any regularization), \textit{w/ SDS(CFG=7.5)}, \textit{w/ SDS(CFG=100)}, and \textit{w/ IPSM(CFG=7.5)} in Tab \ref{tab:comp_sds}. Except for SDS (CFG=7.5), which can provide a limited improvement in structural similarity compared to the Base, the other performances show a downward trend, which is colored by \textcolor{blue}{blue}. However, IPSM can provide considerable improvements in multiple metrics which are colored by \textcolor{red}{red}. It is supposed to be noted that all the experiments of SDS shown in Fig. \ref{fig:sds_trend} and Tab. \ref{tab:comp_sds} are under the same experimental setting. We also present the qualitative comparison of SDS. As shown in Fig. \ref{fig:qualitative_sds} (a), the guidance of SDS will produce the imaginary reconstruction caused by mode deviation when using the diffusion prior directly. This property is reasonable and acceptable in text-to-3D generation tasks, but it fails in specific scene reconstructions limited by sparse views. As shown in Fig. \ref{fig:qualitative_sds} (b), we can observe that SDS will also produce large floaters during optimization, which indicates the characteristic of its training instability since SDS overlooks the inline prior of sparse views and is hard to provide stable guidance towards target mode.

The experiments are conducted 3 times reporting the average results, and use the weight of 2.0 and the VAE encoder same as IPSM for fair comparisons. Since the feature domains of Stable Diffusion and Stable Diffusion Inpainting are identical, using the original VAE of Stable Diffusion shows similar performance, which is reported in the Appendix \ref{sec:appx:different_vae_sds}. We have also analyzed the training instability of SDS additionally in Appendix \ref{sec:appx:training_instablity_sds}. Furthermore, we discuss the effects of using view-conditioned diffusion prior for SDS in Appendix \ref{sec:appx:view_condition_prior}.

\section{Conclusions and Limitations}
\label{sec:conclusion}
In this paper, we start by revisiting the phenomenon where SDS not only fails to improve optimization in sparse-view 3D reconstruction but degrades performance. We present a comprehended analysis of SDS from a mode-seeking perspective. Based on these observations and analyses, we propose Inline Prior Guided Score Matching (IPSM), which utilizes the sparse-view input as the inline prior to rectifying the rendered image distribution. IPSM utilizes the rectified distribution as an intermediate state to decompose the mode-seeking optimization objective of SDS for controlling the optimization direction of mode-seeking to suppress mode deviation. We further propose the pipeline IPSM-Gaussian, which selects 3DGS as the backbone and incorporates IPSM with depth and geometry regularization for boosting IPSM. Experimental results on different public datasets show that our method achieves state-of-the-art reconstruction quality compared to other current methods.

The limitation of our method is that the rectified distribution needs to match the same feature space as the diffusion prior, which restricts the range of inpainting models used for the rectified distribution, thereby limiting the scalability and performance of our method. An alternative improvement could be substituting the pre-trained inpainting models with fine-tuning the diffusion prior like VSD. However, it would further increase the computational complexity of the method. We leave it as our future work.

\section*{Acknowledgement}

This work is partially supported by grants from the National Natural Science Foundation of China under contracts No. 62132002 and No. 62202010, and is also supported by the Fundamental Research Funds for the Central Universities.

\newpage

\bibliographystyle{ieeetr}

% \bibliography{main}

%%%%%%%%%%%%%%%%%%%%%%%%%%%%%%%%%%%%%%%%%%%%%%%%%%%%%%%%%%%%%%

\newpage

\section*{Appendix}

\appendix

\section{Experimental Details}
\label{sec:appx:exp_details}

\subsection{Datasets Details}
\label{sec:appx:dataset_details}

\textbf{LLFF Dataset.} The LLFF dataset \cite{llff} is a forward-facing dataset, which contains 8 challenging scenes. Following FreeNeRF \cite{freenerf} and DNGaussian \cite{dngaussian}, we select every 8th image for testing and evenly sample the remaining images for 3 input views. Following DNGaussian \cite{dngaussian}, we downsample the resolutions of images to $8\times$ for both training and testing. In Tab. \ref{tab:appx:sparse_llff}, we report the level of sparsity for intuitive exhibition. The Original Training Views means the number of training views for the original dense-view NVS, and the Sparsity of 3 Views means the ratio of 3 input sparse views to the Original Training Views.

\begin{table}[H]
\centering
\caption{\textbf{Level of sparsity in the input views of the LLFF dataset}.}\label{tab:appx:sparse_llff}
\setlength{\tabcolsep}{0.1mm}{
\begin{tabular}{c|c|ccccccccc}
\toprule
Dataset & Sparsity & Fer. & Flo. & For. & Hor. & Lea. & Orc. & Roo. & Tre. & AVG \\ \midrule
\multirow{4}{*}{LLFF} 
& Total Views & 20 & 34 & 42 & 62 & 26 & 25 & 41 & 55 & 38.125 \\ 
& Original Training Views & 17 & 29 & 36 & 54 & 22 & 21 & 35 & 48 & 32.750 \\ 
& Test Views & 3 & 5 & 6 & 8 & 4 & 4 & 6 & 7 & 5.375 \\ 
& Sparsity of 3 Views & 17.65\% & 10.34\% & 8.33\% & 5.56\% & 13.64\% & 14.29\% & 8.57\% & 6.25\% & 9.16\% \\ \bottomrule
\end{tabular}
}
\end{table}

\textbf{DTU Dataset.} The DTU dataset \cite{dtu} contains 124 scenes in total. PixelNeRF \cite{pixelnerf} and MVSNeRF \cite{mvsnerf} split the DTU dataset \cite{dtu} into 88 training scenes for pre-training and 15 testing scenes for per-scene fine-tuning. Following RegNeRF \cite{regnerf}, FreeNeRF \cite{freenerf}, and DNGaussian \cite{dngaussian}, we only use the selected 15 testing scenes for optimization. The IDs of testing scenes are: 8, 21, 30, 31, 34, 38, 40, 41, 45, 55, 63, 82, 103, 110, and 114. For each scene optimization with the 3-view setting, the IDs of images served as sparse views for training are 25, 22, and 28. The IDs of images that served as testing novel views for evaluation are 1, 2, 9, 10, 11, 12, 14, 15, 23, 24, 26, 27, 29, 30, 31, 32, 33, 34, 35, 41, 42, 43, 45, 46, 47. Following FreeNeRF \cite{freenerf} and DNGaussian \cite{dngaussian}, all metrics for the evaluation of the DTU dataset are computed with the object mask. Following DNGaussian \cite{dngaussian}, we use the estimated pose which is exactly the same as DNGaussian \cite{dngaussian}. Following RegNeRF \cite{ref-nerf}, we downsample the resolutions of images to $4\times$ for both training and testing.

\subsection{Training Details}
\label{sec:appx:training_details}

\textbf{SfM Initialization.} Following FSGS \cite{fsgs}, we use SfM \cite{sfm} initialization with 3 input sparse views only for the 3D Gaussian points initialization. However, sometimes SfM will fail when using sparse input images. In practice, scan 30 and scan 110 of the DTU dataset cannot extract enough features for initial point cloud prediction, so we only perform random initialization on these two scenes. We perform SfM \cite{sfm} initialization on the remaining scenes of the DTU dataset \cite{dtu} and all scenes of the LLFF dataset \cite{llff}. It is supposed to be noted that SfM \cite{sfm} initialization will significantly improve the final reconstruction quality, so random initialization of these two scenarios will not improve our final performance but must be dealt with due to factual limitations.

\textbf{Gaussian Unpooling.} Following FSGS \cite{fsgs}, we introduce the operation of Gaussian unpooling for filling the spaces uniformly and geometry fitting. The Gaussian unpooling determines whether to add a new Gaussian point by calculating the $K$-nearest neighbor graph structure of the Gaussian point and its corresponding Euclidean distance metric ($K=3$ in practice). The SH coefficients of newly densified Gaussian points are set to 0. In this paper, both experiments of our method, corresponding ablations, and explorations on SDS using different diffusion priors adopt this operation.

\textbf{Gaussian Size Threshold.} The vanilla 3DGS \cite{3dgs} filters out Gaussian points with excessively large sizes, but in the case of sparse views, discarding these large-sized Gaussian points can lead to poor fitting of low-frequency regions during the optimization process. Following FSGS \cite{fsgs} and DNGaussian \cite{dngaussian}, we have eliminated this Gaussian point size filtering operation, which significantly enhances the performance of sparse-view 3D reconstruction.

\textbf{Training Strategy.} Following FSGS \cite{fsgs}, the maximum degree of SH coefficients is set to 3, and we level up the SH degree every 500 iterations. The total training iterations is 10K for all datasets. Following FSGS \cite{fsgs}, we introduce the warm-up period of 500 iterations for the beginning of the pseudo views supervision, \ie the 2K iteration, and we reduce the weight of the depth regularization of seen views to 0.001 after the end of pseudo views supervision, \ie the 9.5K iteration.

\textbf{Background Prior.} Following FreeNeRF \cite{freenerf} and DNGaussian \cite{dngaussian} on the optimization prior based on pixel value (\ie FreeNeRF: white and black background prior; DNGaussian: strategic masking of black backgrounds), we introduce the mask for the white and black background served as an additional prior on the DTU dataset \cite{dtu} for the selection of previous work \cite{regnerf, freenerf, dngaussian} in accurately reconstructing the object-of-interest \cite{regnerf}. Specifically, we mask the values of images that are less than $30/255 \approx 0.1176$ (Following DNGaussian \cite{dngaussian}, the vertical scan rectangles are also introduced to reduce mask of black regions), and larger than $0.99$ for L1 losses of all scenarios on the DTU dataset.

\textbf{Gaussian Points Controlling.} The opacity of Gaussian points would be reset at the 2K iteration. The opacity would not be reset for the following iterations on the LLFF dataset \cite{llff} and the opacity would be reset every 1K iterations for the following iterations on the DTU dataset \cite{dtu} due to the easy over-fitting property associated with large view differences. Besides, the Gaussian points are densified every 100 iterations and pruned every 500 iterations for all datasets.

\textbf{Text Prompts.} For the experiment of score distillation methods, we randomly selected one of the 3 training images and used BLIP \cite{blip} to extract the corresponding text prompts. For fair comparisons, the text prompts corresponding to each scene on all datasets of all score distillation methods relying on text prompts are identical.

\subsection{Hyper-parameters}
\label{sec:appx:hyperparameters}

For the inline prior, the mask threshold $\tau$ is set to 0.3 for IPSM regularization and 0.1 for the geometry consistency regularization, since the latter is at pixel level thus requiring more strict constraints. For the diffusion priors guidance, the weight $\lambda_{\rm{IPSM}}$ of IPSM regularization $\mathcal{L}_{\rm{IPSM}}$ is set to $2.0$ for all datasets, and the parameter $\eta_r$ for controlling $\mathcal{L}_{\rm{IPSM}}^{\mathcal{G}_1}$ and $\mathcal{L}_{\rm{IPSM}}^{\mathcal{G}_2}$ is set to 0.1 for all datasets. The parameter $\eta_d$ for controlling the depth guidance of seen views and pseudo unseen views is set to 0.1 for all datasets. On the LLFF dataset \cite{llff}, the weight $\lambda_{\rm{depth}}$ of depth regularization $\mathcal{L}_{\rm{depth}}$ is set to $0.5$ and the weight $\lambda_{\rm{geo}}$ of the geometry consistency regularization $\mathcal{L}_{\rm{geo}}$ is set to $2.0$. $\lambda_{\rm{ssim}}$ is set to $0.2$ and $\lambda_1 = 1 - \lambda_{\rm{ssim}}$ following 3DGS \cite{3dgs}. On the DTU dataset \cite{dtu}, following DNGaussian \cite{dngaussian}, we reduce $\lambda_1$ to $0.4$ (\ie increase $\lambda_{\rm{ssim}}$ to $0.6$), and at the same time reduce $\lambda_{\rm{depth}}$ and $\lambda_{\rm{geo}}$, both of which are multiplied by $0.1$.

\subsection{Reproduction of Baselines}
\label{sec:appx:baselines}

\textbf{3DGS.} We use the vanilla 3DGS \cite{3dgs} for reproduction on the LLFF \cite{llff} and DTU \cite{dtu} dataset. We do not make any other changes except for the necessary operations to render depth and convert dense views to sparse-view training. In addition, all our experiments use the rasterizer of FSGS \cite{fsgs} to ensure fairness, although this rasterizer has the same function as the rasterizer of 3DGS \cite{3dgs}. Meanwhile, the reported results of 3DGS \cite{3dgs} are also obtained with the SfM \cite{sfm} initialization which is the same as ours.

\textbf{FSGS.} We use the official code of FSGS \cite{fsgs} to reproduce the results on the DTU dataset \cite{dtu}. Since FSGS \cite{fsgs} performed experiments on the LLFF dataset \cite{llff}, we report the results provided by FSGS \cite{fsgs}. Since FSGS \cite{fsgs} does not conduct experiments on the DTU dataset \cite{dtu}, we reproduce it and add the white \& black mask prior to it, which is the same as ours and detailed in Appendix \ref{sec:appx:training_details}. On the DTU dataset \cite{dtu}, we adopt the hyper-parameters of FSGS \cite{fsgs} on the MipNeRF-360 dataset \cite{mipnerf360} (\ie the weight of depth regularization on the pseudo views is $0.03$, the weight of depth regularization on the seen views is $0.05$, and the supervision interval on the pseudo unseen views is $10$) because we observe that the selected hyper-parameters are more suitable for non-forward-facing datasets and can achieve better performance than directly using the hyper-parameters on the LLFF dataset \cite{llff}. The reproduction of FSGS \cite{fsgs} on the DTU dataset \cite{dtu} is also enhanced by the SfM \cite{sfm} initialization same as ours.

\textbf{DNGaussian.} The original DNGaussian \cite{dngaussian} does not use SfM \cite{sfm} initialization. However, directly changing the random initialization to SfM \cite{sfm} initialization without changing the hyper-parameters makes it difficult to provide sufficient performance improvement due to the incompatibility of a series of hyper-parameters such as the learning rate. Therefore, we only report the results provided by DNGaussian \cite{dngaussian} using SfM \cite{sfm} initialization on the LLFF dataset \cite{llff} and do not report the reproduced results of directly using the original random initialization hyper-parameters with SfM \cite{sfm} initialization on the DTU dataset \cite{dtu}.

\subsection{Experimental Environments and Computing Resources}
\label{sec:appx:gpu}

All the experiments are conducted on a single RTX 3090 with CUDA 11.3. The training time of IPSM-Gaussian is about 1 hour on the RTX 3090, which is mainly due to the inference time of the diffusion model itself. For pseudo view supervision from 2K to 9.5K iteration, we need to perform two inferences of the diffusion model in each iteration to calculate $\mathcal{L}_{\rm{IPSM}}^{\mathcal{G}_1}$ and $\mathcal{L}_{\rm{IPSM}}^{\mathcal{G}_2}$.

\section{Additional Experimental Results}
\label{sec:appx:additional_results}

\subsection{Training Instability of SDS}
\label{sec:appx:training_instablity_sds}

In practice, it can be noticed that using SDS directly can produce training instability, which is shown in Tab. \ref{tab:appx:comp_sds_details}. 
Using SDS \cite{sds} causes more performance differences between independent experiments, \ie training instability. The standard deviation of 3 independent experiments using SDS (CFG=7.5) is about 9 times that of IPSM on the SSIM, about 6 times that of IPSM on the LPIPS, and about 3 times that of IPSM on the PSNR. The standard deviation of 3 independent experiments using SDS (CFG=100) is about 3 times that of IPSM on the SSIM, about 3 times that of IPSM on the LPIPS, and about 2 times that of IPSM on the PSNR.
This is because SDS \cite{sds}, as a score distillation technique guided by text-prompt semantics, overlooks the inline priors present in the sparse-view 3D reconstruction task from a limited number of input viewpoints. Owing to the high information entropy inherent in the text, it is hard for SDS to provide stable guidance of diffusion priors towards the target mode during training, leading to instability in the final reconstruction quality.

\begin{table}[t]
\centering
\caption{\textbf{Training instability of SDS}. Detailed data of the reported results in the main manuscript.}
\label{tab:appx:comp_sds_details}
\setlength{\tabcolsep}{1.2mm}{
\begin{tabular}{cccccc}
\toprule
Setting & Experiments & SSIM\textuparrow & LPIPS\textdownarrow & PSNR\textuparrow & AVGE\textdownarrow \\ \midrule
\multirow{4}{*}{Base} & Exp. 1 & 0.619 & 0.260 & 18.89 & 0.128 \\ \cmidrule{2-6}
& Exp. 2 & 0.622 & 0.256 & 18.95 & 0.126 \\ \cmidrule{2-6}
& Exp. 3 & 0.636 & 0.244 & 19.18 & 0.121 \\ \cmidrule{2-6}
& Mean \textpm Std. & 0.625 \textpm 0.008 & 0.254 \textpm 0.007 & 19.00 \textpm 0.12 & 0.125 \textpm 0.003 \\ \midrule
\multirow{4}{*}{w/ SDS(CFG=7.5)} & Exp. 1 & 0.645 & 0.262 & 18.91 & 0.126 \\ \cmidrule{2-6}
& Exp. 2 & 0.637 & 0.282 & 18.29 & 0.136 \\ \cmidrule{2-6}
& Exp. 3 & 0.659 & 0.255 & 19.20 & 0.121 \\ \cmidrule{2-6}
& Mean \textpm Std. & 0.647 \textpm 0.009 & 0.267 \textpm 0.012 & 18.80 \textpm 0.38 & 0.128 \textpm 0.006 \\ \midrule
\multirow{4}{*}{w/ SDS(CFG=100)} & Exp. 1 & 0.571 & 0.375 & 17.20 & 0.167 \\ \cmidrule{2-6}
& Exp. 2 & 0.577 & 0.364 & 17.62 & 0.160 \\ \cmidrule{2-6}
& Exp. 3 & 0.578 & 0.362 & 17.76 & 0.158 \\ \cmidrule{2-6}
& Mean \textpm Std. & 0.576 \textpm 0.003 & 0.367 \textpm 0.006 & 17.53 \textpm 0.24 & 0.162 \textpm 0.004 \\ \midrule
\multirow{4}{*}{w/ IPSM(CFG=7.5)} & Exp. 1 & 0.669 & 0.231 & 19.55 & 0.114 \\ \cmidrule{2-6}
& Exp. 2 & 0.670 & 0.229 & 19.50 & 0.114 \\ \cmidrule{2-6}
& Exp. 3 & 0.672 & 0.227 & 19.76 & 0.111 \\ \cmidrule{2-6}
& Mean \textpm Std. & \textbf{0.670 \textpm 0.001} & \textbf{0.229 \textpm 0.002} & \textbf{19.60 \textpm 0.11} & \textbf{0.113 \textpm 0.001} \\ \bottomrule
\end{tabular}
}
\end{table}

\begin{table}[t]
\centering
\caption{\textbf{Training instability of SDS}. Detailed data of the additional re-conducted results of SDS.}
\label{tab:appx:comp_sds_expand}
\setlength{\tabcolsep}{1.2mm}{
\begin{tabular}{cccccc}
\toprule
Setting & Experiments & SSIM\textuparrow & LPIPS\textdownarrow & PSNR\textuparrow & AVGE\textdownarrow \\ \midrule
\multirow{4}{*}{w/ SDS(CFG=7.5)} & Exp. 1 & 0.643 & 0.272 & 18.78 & 0.129 \\ \cmidrule{2-6}
& Exp. 2 & 0.635 & 0.282 & 18.30 & 0.136 \\ \cmidrule{2-6}
& Exp. 3 & 0.661 & 0.251 & 19.31 & 0.120 \\ \cmidrule{2-6}
& Mean \textpm Std. & 0.647 \textpm 0.011 & 0.268 \textpm 0.013 & 18.80 \textpm 0.41 & 0.128 \textpm 0.007 \\ \midrule
\multirow{4}{*}{w/ SDS(CFG=100)} & Exp. 1 & 0.575 & 0.369 & 17.77 & 0.159 \\ \cmidrule{2-6}
& Exp. 2 & 0.573 & 0.369 & 17.58 & 0.162 \\ \cmidrule{2-6}
& Exp. 3 & 0.583 & 0.361 & 18.03 & 0.154 \\ \cmidrule{2-6}
& Mean \textpm Std. & 0.577 \textpm 0.004 & 0.367 \textpm 0.004 & 17.79 \textpm 0.19 & 0.158 \textpm 0.003 \\ \midrule
\multirow{4}{*}{w/ IPSM(CFG=7.5)} & Exp. 1 & 0.671 & 0.228 & 19.63 & 0.113 \\ \cmidrule{2-6}
& Exp. 2 & 0.673 & 0.227 & 19.68 & 0.112 \\ \cmidrule{2-6}
& Exp. 3 & 0.670 & 0.230 & 19.55 & 0.113 \\ \cmidrule{2-6}
& Mean \textpm Std. & \textbf{0.671 \textpm 0.001} & \textbf{0.228 \textpm 0.001} & \textbf{19.62 \textpm 0.05} & \textbf{0.113 \textpm 0.001} \\ \bottomrule
\end{tabular}
}
\end{table}

To further illustrate the training instability of SDS \cite{sds}, additional experiments of SDS \cite{sds} on the LLFF dataset \cite{llff} are conducted 3 times, which is shown in Tab. \ref{tab:appx:comp_sds_expand}. It is supposed to be noted that the reported results of SDS \cite{sds} in the main manuscript are \textbf{NOT} out of the re-conducted experiments.
In 3 re-conducted experiments, we can still observe the instability exhibited by SDS compared to our method. The standard deviation of 3 independent experiments using SDS (CFG=7.5) is about 11 times that of IPSM on the SSIM, about 13 times that of IPSM on the LPIPS, and about 8 times that of IPSM on the PSNR. The standard deviation of 3 independent experiments using SDS (CFG=100) is about 4 times that of IPSM on the SSIM, about 4 times that of IPSM on the LPIPS, and about 4 times that of IPSM on the PSNR.

\begin{table}[t]
\centering
\caption{\textbf{Comparison to SDS} on the LLFF dataset 3-views setting with different VAE settings.}
\label{tab:appx:comp_sds_samevae}
\setlength{\tabcolsep}{0.7mm}{
\begin{tabular}{ccccccc}
\toprule
Setting & VAE Setting & Experiments & SSIM\textuparrow & LPIPS\textdownarrow & PSNR\textuparrow & AVGE\textdownarrow \\ \midrule
Base & - & Mean \textpm Std. & 0.625 \textpm 0.008 & 0.254 \textpm 0.007 & 19.00 \textpm 0.12 & 0.125 \textpm 0.003 \\ \midrule
\multirow{5}{*}{\begin{tabular}[c]{@{}c@{}}w/ SDS\\(CFG=7.5)\end{tabular}} & Same VAE & Mean \textpm Std. & 0.647 \textpm 0.009 & 0.267 \textpm 0.012 & 18.80 \textpm 0.38 & 0.128 \textpm 0.006 \\ \cmidrule{2-7}
& \multirow{4}{*}{Origin VAE} & Exp. 1 & 0.667 & 0.240 & 19.27 & 0.118 \\ \cmidrule{3-7}
& & Exp. 2 & 0.646 & 0.271 & 18.95 & 0.127 \\ \cmidrule{3-7}
& & Exp. 3 & 0.618 & 0.328 & 17.53 & 0.153 \\ \cmidrule{3-7}
& & Mean \textpm Std. & 0.644 \textpm 0.020 & 0.279 \textpm 0.037 & 18.58 \textpm 0.75 & 0.133 \textpm 0.015 \\ \midrule
\multirow{5}{*}{\begin{tabular}[c]{@{}c@{}}w/ SDS\\(CFG=100)\end{tabular}} & Same VAE & Mean \textpm Std. & 0.576 \textpm 0.003 & 0.367 \textpm 0.006 & 17.53 \textpm 0.24 & 0.162 \textpm 0.004 \\ \cmidrule{2-7}
& \multirow{4}{*}{Origin VAE} & Exp. 1 & 0.578 & 0.364 & 17.97 & 0.156 \\ \cmidrule{3-7}
& & Exp. 2 & 0.568 & 0.374 & 17.55 & 0.163 \\ \cmidrule{3-7}
& & Exp. 3 & 0.567 & 0.377 & 17.52 & 0.164 \\ \cmidrule{3-7}
& & Mean \textpm Std. & 0.571 \textpm 0.005 & 0.372 \textpm 0.005 & 17.68 \textpm 0.21 & 0.161 \textpm 0.004 \\ \midrule
\begin{tabular}[c]{@{}c@{}}w/ IPSM\\(CFG=7.5)\end{tabular} & - & Mean \textpm Std. & \textbf{0.670 \textpm 0.001} & \textbf{0.229 \textpm 0.002} & \textbf{19.60 \textpm 0.11} & \textbf{0.113 \textpm 0.001} \\ \bottomrule
\end{tabular}
}
\end{table}

\begin{table}[t]
\centering
\caption{\textbf{Detials of quantitative comparisons} with other methods.}\label{tab:appx:comp_sota}
\setlength{\tabcolsep}{0.5mm}{
\begin{tabular}{c|c|cccc|cccc}
\toprule
\multirow{2}{*}{Methods} & \multirow{2}{*}{Experiments} & \multicolumn{4}{c|}{LLFF} & \multicolumn{4}{c}{DTU} \\ \cmidrule{3-10} 
& & \multicolumn{1}{l}{SSIM\textuparrow} & \multicolumn{1}{l}{LPIPS\textdownarrow} & \multicolumn{1}{l}{PSNR\textuparrow} & \multicolumn{1}{l|}{AVGE\textdownarrow} & \multicolumn{1}{l}{SSIM\textuparrow} & \multicolumn{1}{l}{LPIPS\textdownarrow} & \multicolumn{1}{l}{PSNR\textuparrow} & \multicolumn{1}{l}{AVGE\textdownarrow} \\ \midrule
\multirow{4}{*}{Ours} & Exp. 1 & 0.703 & 0.207 & 20.55 & 0.100 & 0.857 & 0.119 & 20.13 & 0.076 \\
& Exp. 2 & 0.702 & 0.207 & 20.40 & 0.101 & 0.854 & 0.121 & 19.92 & 0.078 \\
& Exp. 3 & 0.701 & 0.208 & 20.38 & 0.101 & 0.855 & 0.121 & 19.93 & 0.078 \\
& Mean & 0.702 & 0.207 & 20.44 & 0.101 & 0.856 & 0.121 & 19.99 & 0.077 \\ \bottomrule
\end{tabular}
}
\end{table}

\begin{table}[t]
\centering
\caption{\textbf{Details of ablation study} on the LLFF dataset with 3-views setting.}
\label{tab:appx:comp_abla_details}
\setlength{\tabcolsep}{0.8mm}{
\begin{tabular}{cccc|c|cccc}
\toprule
\multicolumn{2}{c}{w/ $\mathcal{L}_{\rm{IPSM}}$} & \multirow{2}{*}{w/ $\mathcal{L}_{\rm{depth}}$} & \multirow{2}{*}{w/ $\mathcal{L}_{\rm{geo}}$} & \multirow{2}{*}{Experiments} & \multirow{2}{*}{SSIM\textuparrow} & \multirow{2}{*}{LPIPS\textdownarrow} & \multirow{2}{*}{PSNR\textuparrow} & \multirow{2}{*}{AVGE\textdownarrow} \\ 
w/ $\mathcal{L}_{\rm{IPSM}}^{\mathcal{G}_1}$ & w/ $\mathcal{L}_{\rm{IPSM}}^{\mathcal{G}_2}$ & & & & &  \\ \midrule
& & & & Mean & 0.625 & 0.254 & 19.00 & 0.125 \\ \midrule
\checkmark & & & & Exp. 1 & 0.638 & 0.243 & 19.23 & 0.120 \\ 
\checkmark & & & & Exp. 2 & 0.641 & 0.242 & 19.24 & 0.120 \\ 
\checkmark & & & & Exp. 3 & 0.631 & 0.249 & 19.18 & 0.122 \\ 
\checkmark & & & & Mean & 0.636 & 0.245 & 19.22 & 0.121 \\ \midrule 
\checkmark & \checkmark & & & Mean & 0.670 & 0.229 & 19.60 & 0.113 \\ \midrule
\checkmark & \checkmark & \checkmark & & Exp. 1 & 0.697 & 0.210 & 20.23 & 0.103 \\ 
\checkmark & \checkmark & \checkmark & & Exp. 2 & 0.699 & 0.211 & 20.22 & 0.103 \\ 
\checkmark & \checkmark & \checkmark & & Exp. 3 & 0.695 & 0.212 & 20.16 & 0.104 \\ 
\checkmark & \checkmark & \checkmark & & Mean & 0.697 & 0.211 & 20.20 & 0.104 \\ \midrule
\checkmark & \checkmark & \checkmark & \checkmark & Mean & 0.702 & 0.207 & 20.44 & 0.101 \\ \bottomrule
\end{tabular}
}
\end{table}

\subsection{SDS with Different VAE}
\label{sec:appx:different_vae_sds}

For a fair comparison, we report the results of SDS on the LLFF dataset \cite{llff} using the VAE same to us, \ie the VAE of Stable Diffusion Inpainting v1-5 \cite{LDM}, in the main manuscript. To demonstrate the reconstruction quality of SDS \cite{sds} in more detail, we report the experimental results of using the original VAE, \ie the VAE of Stable Diffusion v1-5 \cite{LDM}, which is shown in Tab. \ref{tab:appx:comp_sds_samevae}. The experiments employing the original VAE, \ie the VAE of Stable Diffusion v1-5, are also independently repeated thrice. It can be observed that VAE of Stable Diffusion v1-5 and VAE of Stable Diffusion Inpainting v1-5 exhibit nearly identical performances, with the VAE of Stable Diffusion v1-5 (CFG=7.5) even demonstrating greater instability. These experiments further elucidate the mode deviation issue and training instability problem in SDS \cite{sds}.

\subsection{Details of Reported Experimental Results}
\label{sec:appx:details_exp_results}

The details of the reported experimental results in the main manuscript are shown in Tab. \ref{tab:appx:comp_sota} and Tab. \ref{tab:appx:comp_abla_details}. Tab. \ref{tab:appx:comp_sota} shows the details corresponding to the mean and standard deviation of our method as described in Tab. \ref{tab:comp_sota} in the main manuscript, obtained from 3 independent experiments with 3-views setting on the LLFF dataset \cite{llff} and DTU dataset \cite{dtu}, respectively. Tab. \ref{tab:appx:comp_abla_details} shows the details corresponding to the mean and standard deviation as described in Tab. \ref{tab:comp_abla} in the main manuscript, obtained from 3 independent experiments with 3-views setting on the LLFF dataset. Note that the individual results of the first and third row are shown in Tab. \ref{tab:appx:comp_sds_details}. The individual results of the last row are shown in Tab. \ref{tab:appx:comp_sota}.

\subsection{Additional Ablation Results}
\label{sec:appx:additional_ablation}

To supplement more complete experimental results, we provide an additional ablation study using 3 views on the LLFF and DTU dataset in Tab. \ref{tab:appx:addition_llff_ablation} and Tab. \ref{tab:appx:addition_dtu_ablation} respectively. We can see that $\mathcal{L}_{\rm{depth}}$ presents a strong prior for optimization since it directly provides the 3D geometric guidance on 3D representations. Notably, although both $\mathcal{L}_{\rm{geo}}$ and $\mathcal{L}_{\rm{IPSM}}$ utilize re-projection techniques to introduce the 2D visual prior information of the sparse views to promote optimization, $\mathcal{L}_{\rm{IPSM}}$ achieves satisfactory performance comparable to direct 3D guidance of $\mathcal{L}_{\rm{depth}}$ as shown in Tab. \ref{tab:appx:addition_llff_ablation} and Tab. \ref{tab:appx:addition_dtu_ablation}. At the same time, it is difficult for $\mathcal{L}_{\rm{geo}}$ to promote optimization independently without the assistance of other regularizations. 

Besides, in repeated experiments, we also notice that both IPSM and depth regularization can promote the stability of training of 3D Gaussians. As shown in Tab. \ref{tab:appx:addition_llff_ablation}, both IPSM and depth regularization can greatly suppress the fluctuation of reconstruction results in structural similarity and perception evaluation quality, \ie SSIM and LPIPS. However, unlike depth prior, IPSM has a limited suppression effect on the fluctuations of the pixel-level evaluation, \ie PSNR, which is consistent with the randomness of the fluctuations of the baseline as shown in Tab. \ref{tab:appx:addition_llff_ablation} and Tab. \ref{tab:appx:addition_dtu_ablation}. This is because the depth prior participates in optimization throughout the training process (namely [0, 10K] iterations), while IPSM only participates in optimization in [2K, 9.5K] iterations. Due to the significant randomness of 3DGS itself under sparse views \cite{corgs} (especially in more difficult scenarios in DTU compared to LLFF), the optimization of 3DGS itself in the first 2K training iterations may collapse in some scenarios, \eg scan 103, 30, 82, which in turn affects the optimization guidance of the regularization term in subsequent optimizations. Even so, IPSM has a very significant improvement in SSIM and LPIPS compared to $\mathcal{L}_{\rm{geo}}$ which also uses re-projection technology, and is comparable to direct 3D guidance of depth prior as shown in Tab. \ref{tab:appx:addition_dtu_ablation}.

\begin{table}[t]
\centering
\caption{\textbf{Additional ablation study} on the LLFF dataset with 3-views setting.}
\label{tab:appx:addition_llff_ablation}
\setlength{\tabcolsep}{1.5mm}{
\begin{tabular}{c|c|cccc}
\toprule
Setting & Experiments & SSIM\textuparrow & LPIPS\textdownarrow & PSNR\textuparrow & AVGE\textdownarrow \\ \midrule
Base & Mean \textpm Std. & 0.625 \textpm 0.008 & 0.254 \textpm 0.007 & 19.00 \textpm 0.12 & 0.125 \textpm 0.003 \\
\midrule
\multirow{4}{*}{Base + $\mathcal{L}_{\rm{depth}}$} & Exp. 1 & 0.687 & 0.212 & 20.08 & 0.105 \\
& Exp. 2 & 0.690 & 0.210 & 20.18 & 0.104 \\
& Exp. 3 & 0.687 & 0.212 & 20.10 & 0.105 \\
& Mean \textpm Std. & 0.688 \textpm 0.001 & 0.211 \textpm 0.001 & 20.12 \textpm 0.04 & 0.105 \textpm 0.001 \\
\midrule
\multirow{4}{*}{Base + $\mathcal{L}_{\rm{geo}}$} & Exp. 1 & 0.651 & 0.235 & 19.35 & 0.117 \\
& Exp. 2 & 0.643 & 0.240 & 19.14 & 0.120 \\
& Exp. 3 & 0.661 & 0.225 & 19.55 & 0.113 \\
& Mean \textpm Std. & 0.652 \textpm 0.007 & 0.233 \textpm 0.006 & 19.35 \textpm 0.17 & 0.117 \textpm 0.003 \\
\midrule
Base + $\mathcal{L}_{\rm{IPSM}}$ & Mean \textpm Std. & 0.670 \textpm 0.001 & 0.229 \textpm 0.002 & 19.60 \textpm 0.11 & 0.113 \textpm 0.001 \\
\midrule
Ours & Mean \textpm Std. & 0.702 \textpm 0.001 & 0.207 \textpm 0.001 & 20.44 \textpm 0.08 & 0.101 \textpm 0.001 \\
\bottomrule
\end{tabular}
}
\end{table}

\begin{table}[t]
\centering
\caption{\textbf{Additional ablation study} on the DTU dataset with 3-views setting.}
\label{tab:appx:addition_dtu_ablation}
\setlength{\tabcolsep}{1.5mm}{
\begin{tabular}{c|c|cccc}
\toprule
Setting & Experiments & SSIM\textuparrow & LPIPS\textdownarrow & PSNR\textuparrow & AVGE\textdownarrow \\ \midrule
\multirow{4}{*}{Base} & Exp. 1 & 0.836 & 0.134 & 19.11 & 0.087 \\
& Exp. 2 & 0.836 & 0.135 & 18.86 & 0.089 \\
& Exp. 3 & 0.837 & 0.134 & 19.39 & 0.085 \\
& Mean \textpm Std. & 0.836 \textpm 0.001 & 0.134 \textpm 0.001 & 19.12 \textpm 0.22 & 0.087 \textpm 0.002 \\
\midrule
\multirow{4}{*}{Base + $\mathcal{L}_{\rm{depth}}$} & Exp. 1 & 0.849 & 0.122 & 19.77 & 0.079 \\
& Exp. 2 & 0.853 & 0.121 & 19.92 & 0.078 \\
& Exp. 3 & 0.852 & 0.121 & 19.77 & 0.079 \\
& Mean \textpm Std. & 0.851 \textpm 0.001 & 0.122 \textpm 0.001 & 19.82 \textpm 0.07 & 0.079 \textpm 0.001 \\
\midrule
\multirow{4}{*}{Base + $\mathcal{L}_{\rm{geo}}$} & Exp. 1 & 0.835 & 0.135 & 19.28 & 0.086 \\
& Exp. 2 & 0.833 & 0.137 & 18.86 & 0.090 \\
& Exp. 3 & 0.837 & 0.134 & 19.41 & 0.085 \\
& Mean \textpm Std. & 0.835 \textpm 0.001 & 0.135 \textpm 0.001 & 19.18 \textpm 0.23 & 0.087 \textpm 0.002 \\
\midrule
\multirow{4}{*}{Base + $\mathcal{L}_{\rm{IPSM}}$} & Exp. 1 & 0.853 & 0.122 & 19.67 & 0.080 \\
& Exp. 2 & 0.852 & 0.123 & 19.80 & 0.079 \\
& Exp. 3 & 0.850 & 0.125 & 19.34 & 0.083 \\
& Mean \textpm Std. & 0.852 \textpm 0.001 & 0.123 \textpm 0.001 & 19.60 \textpm 0.19 & 0.080 \textpm 0.002 \\
\midrule
Ours & Mean \textpm Std. & 0.856 \textpm 0.001 & 0.121 \textpm 0.001 & 19.99 \textpm 0.10 & 0.077 \textpm 0.001 \\
\bottomrule
\end{tabular}
}
\end{table}

\subsection{Additional Experiments with Different Input Views}
\label{sec:appx:different_views}

\textbf{More input views}. Experimental results using more input views can further explore the robustness of our method when working with sparse views. We provide additional experimental results under 6 and 9 input views on the LLFF dataset in Tab. \ref{tab:appx:LLFF_6_views} and Tab. \ref{tab:appx:LLFF_9_views} respectively. Notably, our method uses exactly the same parameters as the LLFF dataset with 3 views for training. For the \textbf{6 input views}, as shown in Tab. \ref{tab:appx:LLFF_6_views}, we achieve an improvement of $11.18\%$ on LPIPS compared to ReconFusion \cite{reconfusion}. It is supposed to be noted that ReconFusion \cite{reconfusion} requires additional computational resources for pre-training an encoder with external data as we demonstrated in the main manuscript. Excluding methods that require additional resources for pre-training, our method achieves improvements of $7.94\%$, $8.34\%$, $31.82\%$, $30.68\%$ on PSNR, SSIM, LPIPS, and AVGE respectively, compared to DNGaussian \cite{dngaussian}, which is the state-of-the-art method based on the 3DGS \cite{3dgs}. For the \textbf{9 input views}, similar to the experimental results of 6 input views, our method still outperforms all state-of-the-art methods on SSIM, LPIPS, and AVGE scores and achieves comparable results on PSNR. As shown in Tab. \ref{tab:appx:LLFF_9_views}, compared to 3DGS-based DNGaussian \cite{dngaussian}, we achieve improvements of $8.46\%$, $8.50\%$, $38.33\%$, $33.77\%$ on PSNR, SSIM, LPIPS, and AVGE respectively.

\begin{table}[t]
\centering
\caption{\textbf{Quantitative comparisons with 6 input views} on the LLFF dataset.}
\label{tab:appx:LLFF_6_views}
\setlength{\tabcolsep}{0.5mm}{
\begin{tabular}{ccccccc}
\toprule
Method & Pretrain & Experiments & SSIM\textuparrow & LPIPS\textdownarrow & PSNR\textuparrow & AVGE\textdownarrow \\ \midrule
Zip-NeRF * & - & - & 0.764 & 0.221 & 20.71 & 0.097 \\ \midrule
RegNeRF * \cite{regnerf} & - & - & 0.760 & 0.243 & 23.09 & 0.084 \\ \midrule
DiffusioNeRF * & \checkmark & - & 0.775 & 0.235 & 23.60 & 0.079 \\ \midrule
FreeNeRF * \cite{freenerf} & - & - & 0.773 & 0.232 & 23.72 & 0.078 \\ \midrule
SimpleNeRF * \cite{simplenerf} & - & - & 0.737 & 0.296 & 23.05 & 0.091 \\ \midrule
ReconFusion * \cite{reconfusion} & \checkmark & - & 0.815 & 0.152 & \textbf{24.25} & 0.063 \\ \midrule
3DGS \# \cite{3dgs} & - & - & 0.699 & 0.226 & 20.63 & 0.108 \\ \midrule
DNGaussian \# \cite{dngaussian} & - & - & 0.755 & 0.198 & 22.18 & 0.088 \\ \midrule
\multirow{4}{*}{Ours} & - & Exp. 1 & 0.818 & 0.135 & 23.98 & 0.061 \\ \cmidrule{2-7}
& - & Exp. 2 & 0.819 & 0.135 & 23.95 & 0.061 \\ \cmidrule{2-7}
& - & Exp. 3 & 0.818 & 0.135 & 23.91 & 0.062 \\ \cmidrule{2-7}
& - & Mean \textpm Std. & \textbf{0.818 \textpm 0.001} & \textbf{0.135 \textpm 0.001} & 23.94 \textpm 0.03 & \textbf{0.061 \textpm 0.001} \\ \midrule
\end{tabular}
}
\begin{tablenotes}
    \item[*] *: results reported in ReconFusion \cite{reconfusion}.
    \item[\#] \#: results reported in DNGaussian \cite{dngaussian}.
\end{tablenotes}
\end{table}

\begin{table}[t]
\centering
\caption{\textbf{Quantitative comparisons with 9 input views} on the LLFF dataset.}
\label{tab:appx:LLFF_9_views}
\setlength{\tabcolsep}{0.5mm}{
\begin{tabular}{ccccccc}
\toprule
Method & Pretrain & Experiments & SSIM\textuparrow & LPIPS\textdownarrow & PSNR\textuparrow & AVGE\textdownarrow \\ \midrule
Zip-NeRF * & - & - & 0.830 & 0.166 & 23.63 & 0.067 \\ \midrule
RegNeRF * \cite{regnerf} & - & - & 0.820 & 0.196 & 24.84 & 0.065 \\ \midrule
DiffusioNeRF * & \checkmark & - & 0.807 & 0.216 & 24.62 & 0.069 \\ \midrule
FreeNeRF * \cite{freenerf} & - & - & 0.820 & 0.193 & 25.12 & 0.063 \\ \midrule
SimpleNeRF * \cite{simplenerf} & - & - & 0.762 & 0.286 & 23.98 & 0.082 \\ \midrule
ReconFusion * \cite{reconfusion} & \checkmark & - & 0.848 & 0.134 & \textbf{25.21} & 0.054 \\ \midrule
3DGS \# \cite{3dgs} & - & - & 0.697 & 0.230 & 20.44 & 0.108 \\ \midrule
DNGaussian \# \cite{dngaussian} & - & - & 0.788 & 0.180 & 23.17 & 0.077 \\ \midrule
\multirow{4}{*}{Ours} & - & Exp. 1 & 0.854 & 0.113 & 25.02 & 0.051 \\ \cmidrule{2-7}
& - & Exp. 2 & 0.856 & 0.111 & 25.20 & 0.050 \\ \cmidrule{2-7}
& - & Exp. 3 & 0.856 & 0.110 & 25.19 & 0.050 \\ \cmidrule{2-7}
& - & Mean \textpm Std. & \textbf{0.855 \textpm 0.001} & \textbf{0.111 \textpm 0.001} & 25.13 \textpm 0.08 & \textbf{0.051 \textpm 0.001} \\ \midrule
\end{tabular}
}
\begin{tablenotes}
    \item[*] *: results reported in ReconFusion \cite{reconfusion}.
    \item[\#] \#: results reported in DNGaussian \cite{dngaussian}.
\end{tablenotes}
\end{table}

\textbf{Less input views}. To evaluate extreme circumstances, \eg opposite views and extrapolation scenarios, we construct corresponding data and conduct experiments with the state-of-the-art method DNGaussian \cite{dngaussian}. For the \textbf{two opposite input views}, we select 2 opposite views of each scene on the MipNeRF-360 dataset, i.e. the IDs of training views of each scene: 2, 26 of bicycle; 22, 151 of bonsai; 57, 185 of counter; 1, 57 of garden; 14, 171 of kitchen; 2, 79 of room; 26, 34 of stump. The test views are selected every 8th image following Mip-NeRF. The quantitative comparisons with state-of-the-art method DNGaussian \cite{dngaussian} are shown in Tab. \ref{tab:appx:comp_mip360_2_views}. It can be seen that our method outperforms DNGaussian \cite{dngaussian} and our model achieves improvements of $21.90\%$, $18.86\%$ on average PSNR and AVGE scores respectively. For the \textbf{extrapolation scenarios}, We select 2 views on 0 and 90 degrees of each scene on the MipNeRF-360 dataset, i.e. IDs: 2, 14 of bicycle; 22, 248 of bonsai; 57, 145 of counter; 1, 15 of garden; 14, 37 of kitchen; 2, 291 of room; 26, 28 of stump. The test views are selected on the 180 degrees, i.e. IDs: 26 of bicycle; 151 of bonsai; 185 of counter; 57 of garden; 171 of kitchen; 79 of room; 34 of stump. The quantitative results similar to opposite views are shown in Tab. \ref{tab:appx:comp_mip360_2_views}. It can be seen that our method outperforms the state-of-the-art method DNGaussian \cite{dngaussian} and our model achieves improvements of $27.27\%$, $22.57\%$ on average PSNR and AVGE scores respectively. We can notice that although our method is improved compared to DNGaussian, in fact, current sparse-view reconstruction methods (including our method) cannot successfully reconstruct extreme cases. We leave it as our future work.

\begin{table}[t]
\centering
\caption{\textbf{Quantitative comparisons} with 2 views on the MipNeRF-360 dataset.}\label{tab:appx:comp_mip360_2_views}
\setlength{\tabcolsep}{0.5mm}{
\begin{tabular}{c|c|c|cccc}
\toprule
Metric & Methods & Experiments & SSIM\textuparrow & LPIPS\textdownarrow & PSNR\textuparrow & AVGE\textdownarrow \\ \midrule
\multirow{8}{*}{\begin{tabular}[c]{@{}c@{}}Opposite\\Views\end{tabular}} & \multirow{4}{*}{DNGaussian} & Exp. 1 & 0.142 & 0.705 & 10.53 & 0.387 \\
& & Exp. 2 & 0.141 & 0.704 & 10.49 & 0.388 \\
& & Exp. 3 & 0.142 & 0.705 & 10.49 & 0.388 \\
& & Mean \textpm Std. & 0.142 \textpm 0.001 & 0.705 \textpm 0.001 & 10.50 \textpm 0.02 & 0.387 \textpm 0.001 \\ \cline{2-7}
& \multirow{4}{*}{Ours} & Exp. 1 & 0.243 & 0.677 & 12.85 & 0.313 \\ 
& & Exp. 2 & 0.245 & 0.675 & 12.78 & 0.314 \\
& & Exp. 3 & 0.242 & 0.678 & 12.77 & 0.315 \\
& & Mean \textpm Std. & \textbf{0.243 \textpm 0.001} & \textbf{0.677 \textpm 0.001} & \textbf{12.80 \textpm 0.04} & \textbf{0.314 \textpm 0.001} \\ \hline
\multirow{8}{*}{\begin{tabular}[c]{@{}c@{}}Extrapolation\\Scenarios\end{tabular}} & \multirow{4}{*}{DNGaussian} & Exp. 1 & 0.075 & 0.734 & 9.89 & 0.417 \\
& & Exp. 2 & 0.063 & 0.739 & 9.67 & 0.426 \\
& & Exp. 3 & 0.081 & 0.736 & 9.81 & 0.419 \\
& & Mean \textpm Std. & 0.073 \textpm 0.007 & 0.736 \textpm 0.002 & 9.79 \textpm 0.09 & 0.421 \textpm 0.004 \\ \cline{2-7}
& \multirow{4}{*}{Ours} & Exp. 1 & 0.267 & 0.707 & 12.61 & 0.322 \\ 
& & Exp. 2 & 0.266 & 0.711 & 12.44 & 0.326 \\
& & Exp. 3 & 0.258 & 0.712 & 12.33 & 0.330 \\
& & Mean \textpm Std. & \textbf{0.264 \textpm 0.004} & \textbf{0.710 \textpm 0.002} & \textbf{12.46 \textpm 0.11} & \textbf{0.326 \textpm 0.003} \\
\bottomrule
\end{tabular}
}
\end{table}

\subsection{Additional Evaluation and Discussion of View-conditioned Diffusion Priors}
\label{sec:appx:view_condition_prior}

It is worth noting that the \textbf{\textit{SDS mentioned before are all based on the 2D diffusion priors}}. A natural idea is that we can use the 3D diffusion prior with the vanilla SDS to promote sparse-view 3D reconstruction without designing a complex method to extract 3D visual knowledge from the 2D diffusion prior. In this section, we discuss using view-conditioned 3D diffusion priors with SDS to improve the reconstruction quality under sparse views. We conduct experiments on the LLFF dataset with 3 views using view-conditioned 3D diffusion priors to evaluate their visual guidance of them. Specifically, we use the 3D prior, \ie Zero-1-to-3 \cite{zero123} and ZeroNVS \cite{zeronvs}, and their default CFG to optimize the 3D scene under sparse views through the vanilla SDS. We also use the same backbone and weights as IPSM. Besides, we explore the effect of warmup operation for the SDS regularization of 3D priors.

As shown in Tab. \ref{tab:appx:view_condition_prior}, the first three rows and the last row are the experimental results mentioned before. The fourth line shows the result of using the Inpainting Stable Diffusion model (ISD) with inline priors to assist SDS, which is actually the ablation result of $\mathcal{L}_{\rm{IPSM}}^{\mathcal{G}_1}$ in the ablation experiment shown in Tab. \ref{tab:comp_abla}. We can notice that both Zero-1-to-3 and ZeroNVS can only provide limited visual guidance and may even hinder reconstruction compared to the Baseline. 
Besides, using ZeroNVS \cite{zeronvs} is superior compared to using Zero-1-to-3 \cite{zero123} since the former utilizes 3D annotated scene data for fine-tuning while Zero-1-to-3 only uses 3D objects dataset for fine-tuning. However, although ZeroNVS \cite{zeronvs} as 3D prior can achieve stunning results in single-view reconstruction for inferring 3D structure from an unlabeled 2D image \cite{zeronvs}, it still cannot boost the sparse-view reconstruction quality as IPSM since the ZeroNVS guidance does not exploit inline priors for sparse views which is different from the single-view setting. 

Currently, 3D diffusion priors already have a certain ability to represent the 3D world. However, as reported experimental results in Tab. \ref{tab:appx:view_condition_prior}, 3D diffusion priors still cannot provide a significant boost on different 3D scene datasets, since the scarcity of 3D annotation data used to fine-tune 3D diffusion priors exists. Specifically, ZeroNVS \cite{zeronvs} fine-tuned on a mixture million-level dataset consisting of CO3D \cite{co3d}, ACID \cite{acid_1, acid_2}, and RealEstate10K \cite{realestate}. But, Stable Diffusion \cite{LDM} and its inpainting version are trained on billion-level LAION-5B \cite{laion5b}. With the additional conducted experiments, we notice that there is still an objective fact that 3D training data for 3D diffusion models is scarce. How to efficiently construct high-fidelity 3D data, or how to use 2D data knowledge to complement the training of 3D diffusion prior remains a core challenge in this field.

\begin{table}[t]
\centering
\caption{\textbf{Quantitative experimental results} using view-conditioned diffusion priors on the LLFF dataset with 3-views setting.}
\label{tab:appx:view_condition_prior}
\setlength{\tabcolsep}{1.0mm}{
\begin{tabular}{c|c|cccc}
\toprule
Setting & Experiments & SSIM\textuparrow & LPIPS\textdownarrow & PSNR\textuparrow & AVGE\textdownarrow \\ \midrule
Base & Mean \textpm Std. & 0.625 \textpm 0.008 & 0.254 \textpm 0.007 & 19.00 \textpm 0.12 & 0.125 \textpm 0.003 \\
\midrule
SD, CFG=7.5 & Mean \textpm Std. & 0.647 \textpm 0.009 & 0.267 \textpm 0.012 & 18.80 \textpm 0.38 & 0.128 \textpm 0.006 \\
\midrule
SD, CFG=100 & Mean \textpm Std. & 0.576 \textpm 0.003 & 0.367 \textpm 0.006 & 17.53 \textpm 0.24 & 0.162 \textpm 0.004 \\
\midrule
ISD(\ie $\mathcal{L}_{\rm{IPSM}}^{\mathcal{G}_1}$), CFG=7.5 & Mean \textpm Std. & 0.636 \textpm 0.004 & 0.245 \textpm 0.003 & 19.22 \textpm 0.02 & 0.121 \textpm 0.001 \\
\midrule
\multirow{4}{*}{\begin{tabular}[c]{@{}c@{}}Zero-1-to-3, CFG=3.0\end{tabular}} & Exp. 1 & 0.566 & 0.361 & 17.65 & 0.160 \\
& Exp. 2 & 0.576 & 0.354 & 17.70 & 0.158 \\
& Exp. 3 & 0.577 & 0.351 & 18.00 & 0.153 \\
& Mean \textpm Std. & 0.573 \textpm 0.005 & 0.355 \textpm 0.004 & 17.78 \textpm 0.15 & 0.157 \textpm 0.003 \\
\midrule
\multirow{4}{*}{\begin{tabular}[c]{@{}c@{}}Zero-1-to-3, CFG=3.0\\ w/ WarmUp\end{tabular}} & Exp. 1 & 0.584 & 0.344 & 17.81 & 0.154 \\
& Exp. 2 & 0.576 & 0.349 & 17.87 & 0.155 \\
& Exp. 3 & 0.575 & 0.361 & 17.79 & 0.158 \\
& Mean \textpm Std. & 0.578 \textpm 0.004 & 0.352 \textpm 0.007 & 17.82 \textpm 0.03 & 0.156 \textpm 0.001 \\
\midrule
\multirow{4}{*}{\begin{tabular}[c]{@{}c@{}}ZeroNVS, CFG=7.5\end{tabular}} & Exp. 1 & 0.639 & 0.289 & 19.12 & 0.129 \\
& Exp. 2 & 0.633 & 0.292 & 19.15 & 0.129 \\
& Exp. 3 & 0.641 & 0.286 & 19.40 & 0.125 \\
& Mean \textpm Std. & 0.638 \textpm 0.003 & 0.289 \textpm 0.003 & 19.22 \textpm 0.12 & 0.128 \textpm 0.002 \\
\midrule
\multirow{4}{*}{\begin{tabular}[c]{@{}c@{}}ZeroNVS, CFG=7.5\\ w/ WarmUp\end{tabular}} & Exp. 1 & 0.647 & 0.281 & 19.22 & 0.126 \\
& Exp. 2 & 0.643 & 0.283 & 19.29 & 0.126 \\
& Exp. 3 & 0.644 & 0.282 & 19.30 & 0.126 \\
& Mean \textpm Std. & 0.645 \textpm 0.001 & 0.282 \textpm 0.001 & 19.27 \textpm 0.04 & 0.126 \textpm 0.001 \\
\midrule
IPSM(Ours), CFG=7.5 & Mean \textpm Std. & \textbf{0.670 \textpm 0.001} & \textbf{0.229 \textpm 0.002} & \textbf{19.60 \textpm 0.11} & \textbf{0.113 \textpm 0.001} \\
\bottomrule
\end{tabular}
}
\end{table}

\subsection{Intuitive Explanation of Inline Priors}
\label{sec:appx:vis_inline}

\begin{figure}[t]
    \centering 
    \includegraphics[width=1.0\linewidth]{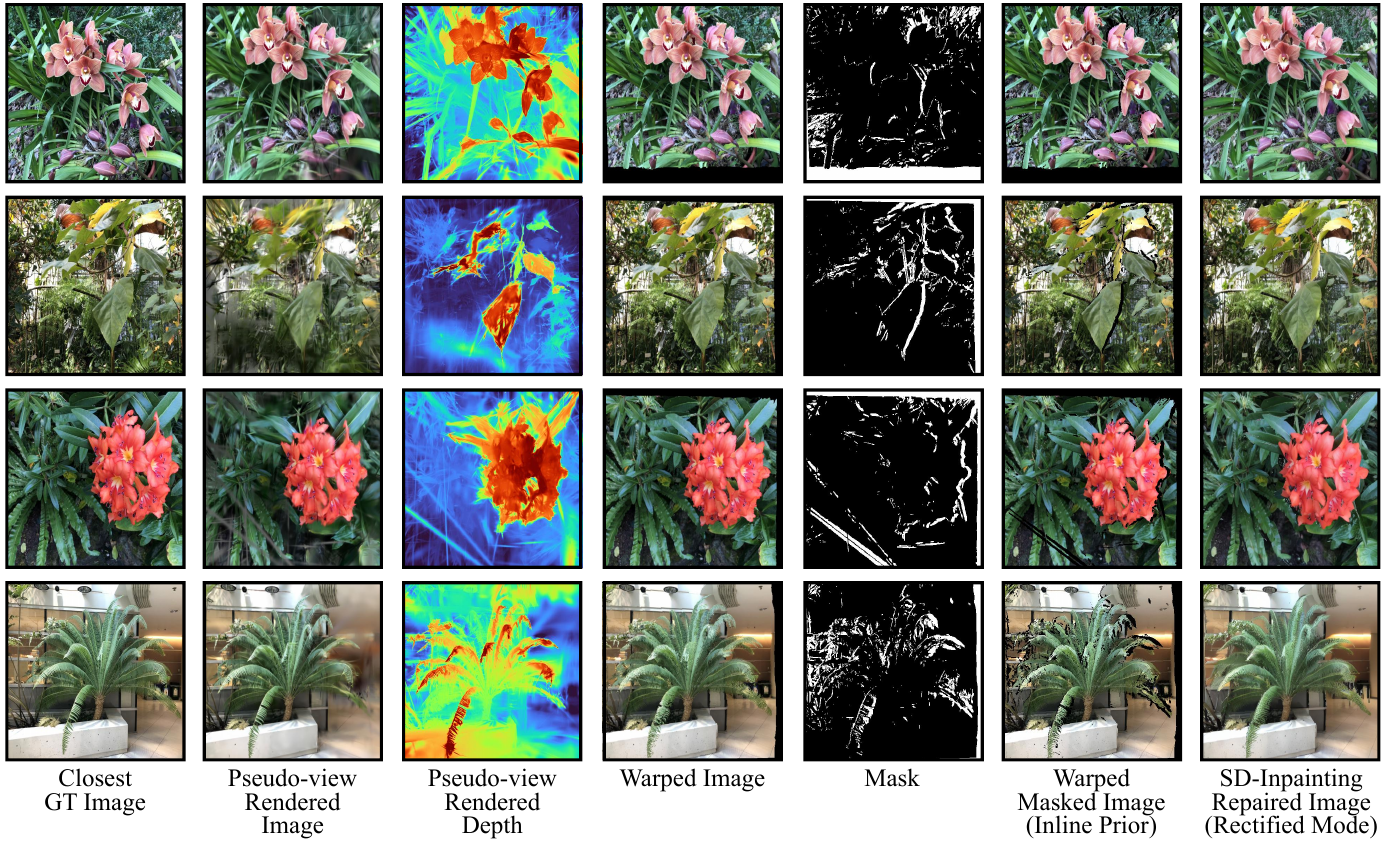}
    \caption{\textbf{Intuitive explanation of the inline priors}.} 
    \label{fig:appx:inline_prior}
\end{figure}

To visually demonstrate the effect of inline priors for rectification on the diffusion prior more intuitively, we show the inline priors along with their associated visual content in Fig. \ref{fig:appx:inline_prior} as a intuitive supplement to our motivation. Note that we choose a relatively tight depth error threshold to better illustrate the potential of the rectified distribution. 
The first column shows the input sparse seen-view image; the second column includes the rendering images of the pseudo unseen view which is sampled around the seen view (Eq. \ref{eq:render_rgb}); the third column presents the rendering depths corresponding to the pseudo view (Eq. \ref{eq:render_depth}); the fourth column consists of the warped images obtained based on the pose transformation relationships with the seen-view image, pseudo-view rendering depth (Eq. \ref{eq:inverse_warp}); the fifth column depicts masks derived from the depth differences (Eq. \ref{eq:warp_mask}); the sixth column displays the masked warped images, known as inline priors, which integrate visual inline information from the seen view to the pseudo unseen views, thereby laying the foundation for subsequent rectification of the diffusion prior; and the seventh column intuitively exhibits images obtained through 25-step sampling using noise-added rendering images served as latents and inline priors served as conditions with Stable Diffusion Inpainting v1-5 \cite{LDM}, representing the rectified mode of the corresponding scenes in the rectified distribution.

\begin{figure}[t]
    \centering 
    \includegraphics[width=1.0\linewidth]{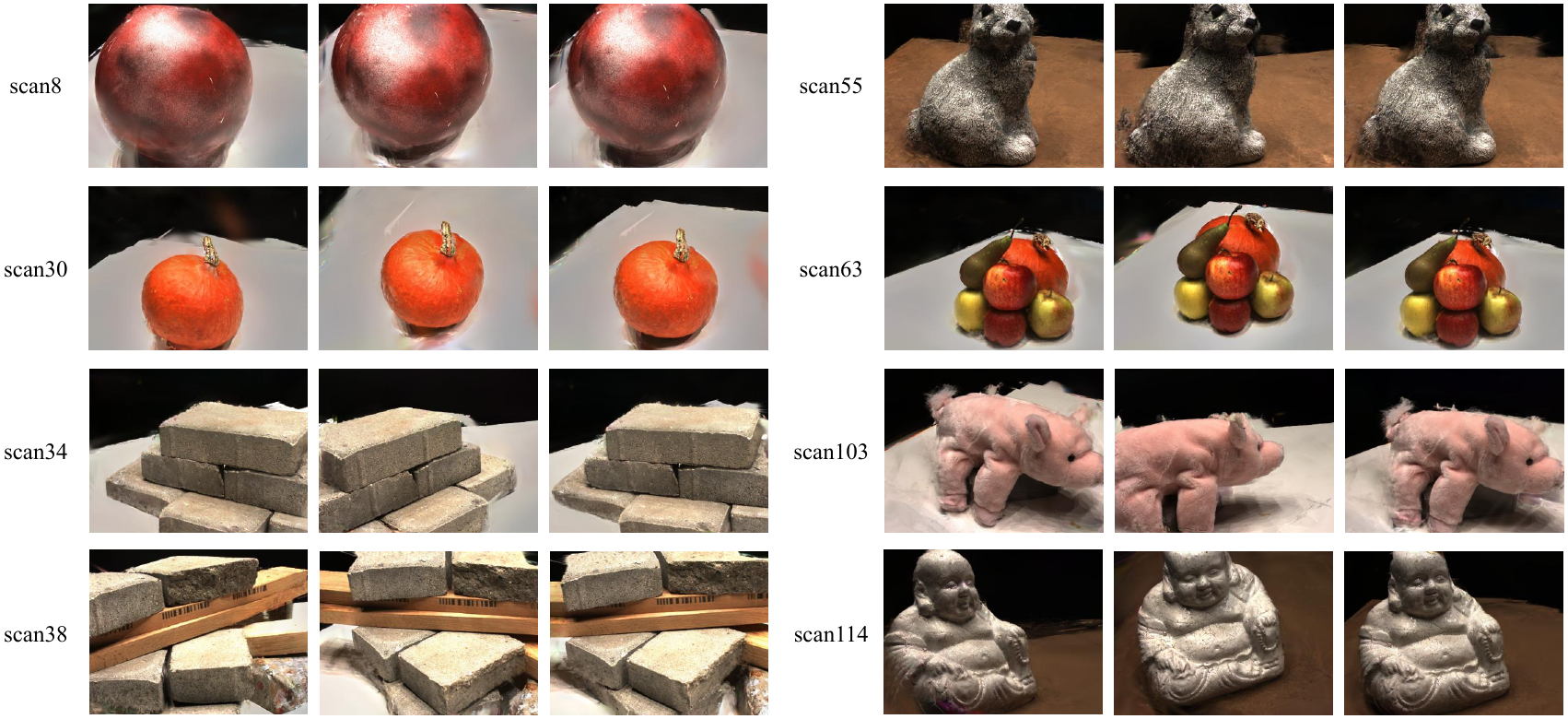} 
    \caption{\textbf{Examples of novel view synthesis from our method with 3 input views on the DTU dataset}.} 
    \label{fig:appx:qualitative_dtu}
\end{figure}

\begin{figure}[t]
    \centering 
    \includegraphics[width=1.0\linewidth]{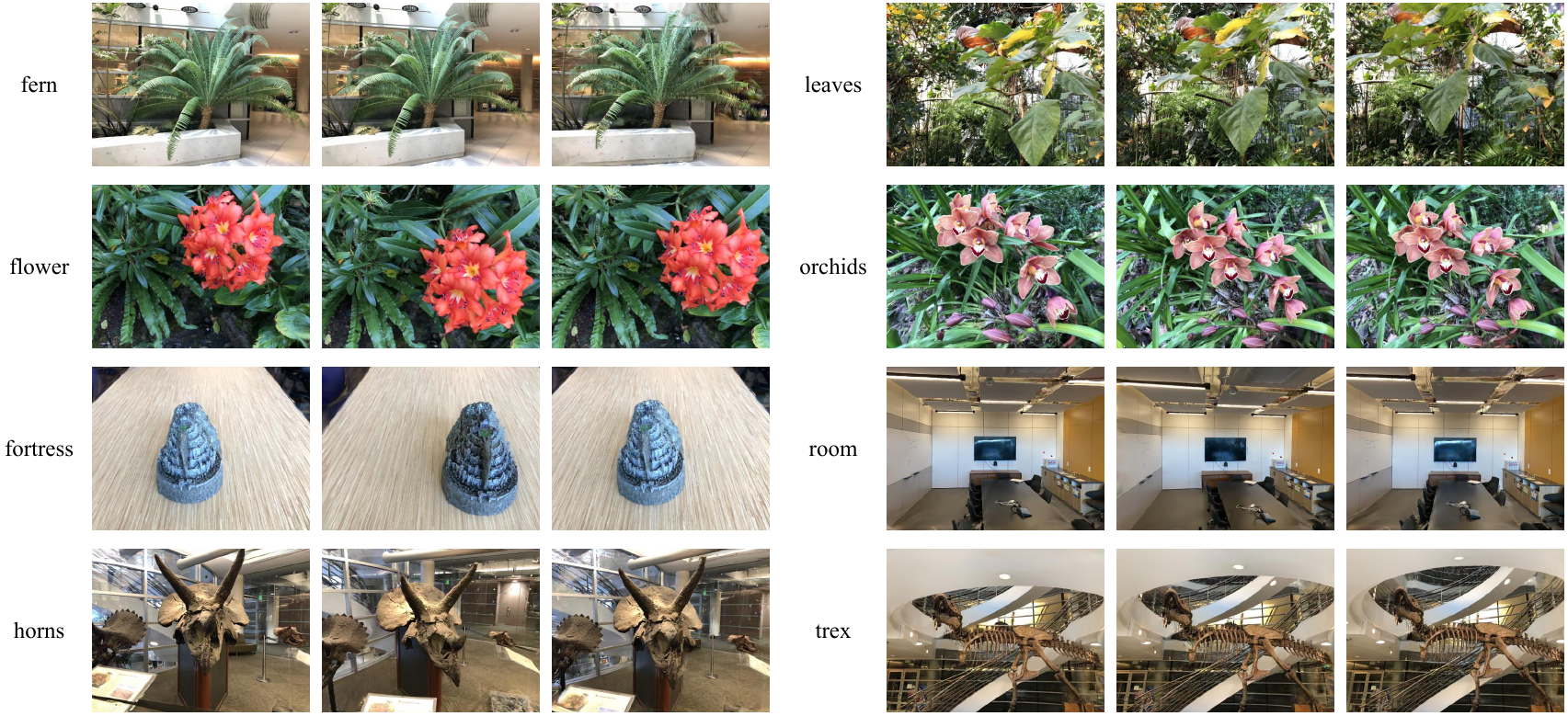} 
    \caption{\textbf{Examples of novel view synthesis from our method with 3 input views on the LLFF dataset}.} 
    \label{fig:appx:qualitative_llff}
\end{figure}

\subsection{More Qualitative Results}
\label{sec:appx:qualitative_results}

We present additional examples of rendered images in the test set shown in Fig. \ref{fig:appx:qualitative_dtu} and Fig. \ref{fig:appx:qualitative_llff}. The examples of rendering results are obtained from the DTU dataset \cite{dtu} and the LLFF dataset \cite{llff} with 3 training views. More qualitative results can be found in our \textcolor{blue}{\textbf{supplementary video}}.  

\end{document}